\documentclass[fleqn,10pt]{wlscirep}

\usepackage[utf8]{inputenc}
\usepackage[T1]{fontenc}

\usepackage{microtype}
\usepackage{graphicx}
\usepackage{subfigure}
\usepackage{booktabs} 
\usepackage{hyperref}
\usepackage[font=small,justification=justified]{caption}

\usepackage{amsmath,amssymb}
\usepackage{bm}
\usepackage{adjustbox}
\usepackage{hhline}
\usepackage{boldline}
\usepackage{array,multirow,graphicx}
\usepackage{float}

\usepackage[sorting=none,defernumbers=true,backend=bibtex,style=nature, autocite=superscript]{biblatex}

\fancypagestyle{firstpage}{          
\lhead{}%
\chead{}%
\rhead{}%
\lfoot{}%
\cfoot{Preprint Version. Published November, 2020, Nature Machine Intelligence. DOI: \href{https://doi.org/10.1038/s42256-020-00255-1}{10.1038/s42256-020-00255-1} }%
\rfoot{}%
}

\title{Evolutionary Training and Abstraction Yields Algorithmic Generalization of Neural Computers}

\author[1,*]{Daniel Tanneberg}
\author[2,1]{Elmar Rueckert}
\author[1,3]{Jan Peters}
\affil[1]{Intelligent Autonomous Systems, Technische Universit\"at Darmstadt, Darmstadt, Germany}
\affil[2]{Institute for Robotics and Cognitive Systems, Universit\"at zu L\"ubeck, L\"ubeck, Germany}
\affil[3]{Robot Learning Group, Max-Planck Institute for Intelligent Systems, T\"ubingen, Germany}

\affil[*]{daniel@robot-learning.de}

\newcommand{\figureNHC}{
\begin{figure*}[!t]
	\centering  
	\includegraphics[width=0.99\textwidth]{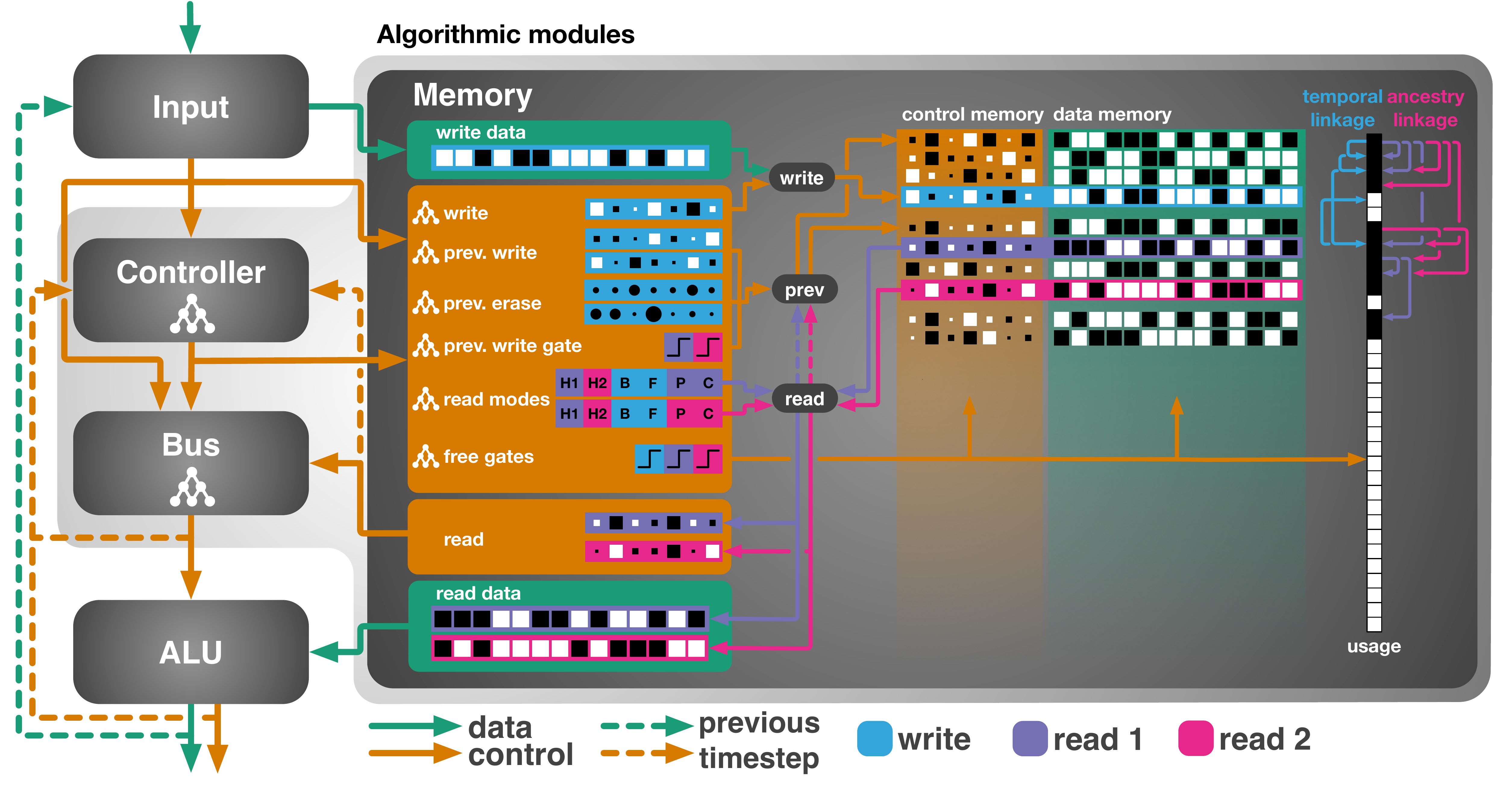}
	\caption{
	\textbf{The neural harvard computer architecture.}
	Information flow is divided into \textit{data} (green) and \textit{control} (orange) streams. 
	The modules inside the light grey area -- the controller, the memory and the bus -- are learning the algorithmic solution on the control stream, whereas the data modules are either learned beforehand or hand-designed.
	The algorithmic solution operates solely on the control stream to steer the data access and manipulation, whereas the learning signal can be provided on any connection in the architecture (data or control) due to the evolution based training.
	Inside the memory module the learnable interfaces which control the data access to the two memory matrices are shown.
	Sign and magnitude of vectors are shown as the colour and size of the boxes and circles.
	} 
	\label{fig:nhc} 
\end{figure*}
}

\newcommand{\figureLearning}{
\begin{figure*}[!t]
	\centering  
	\includegraphics[width=0.99\textwidth]{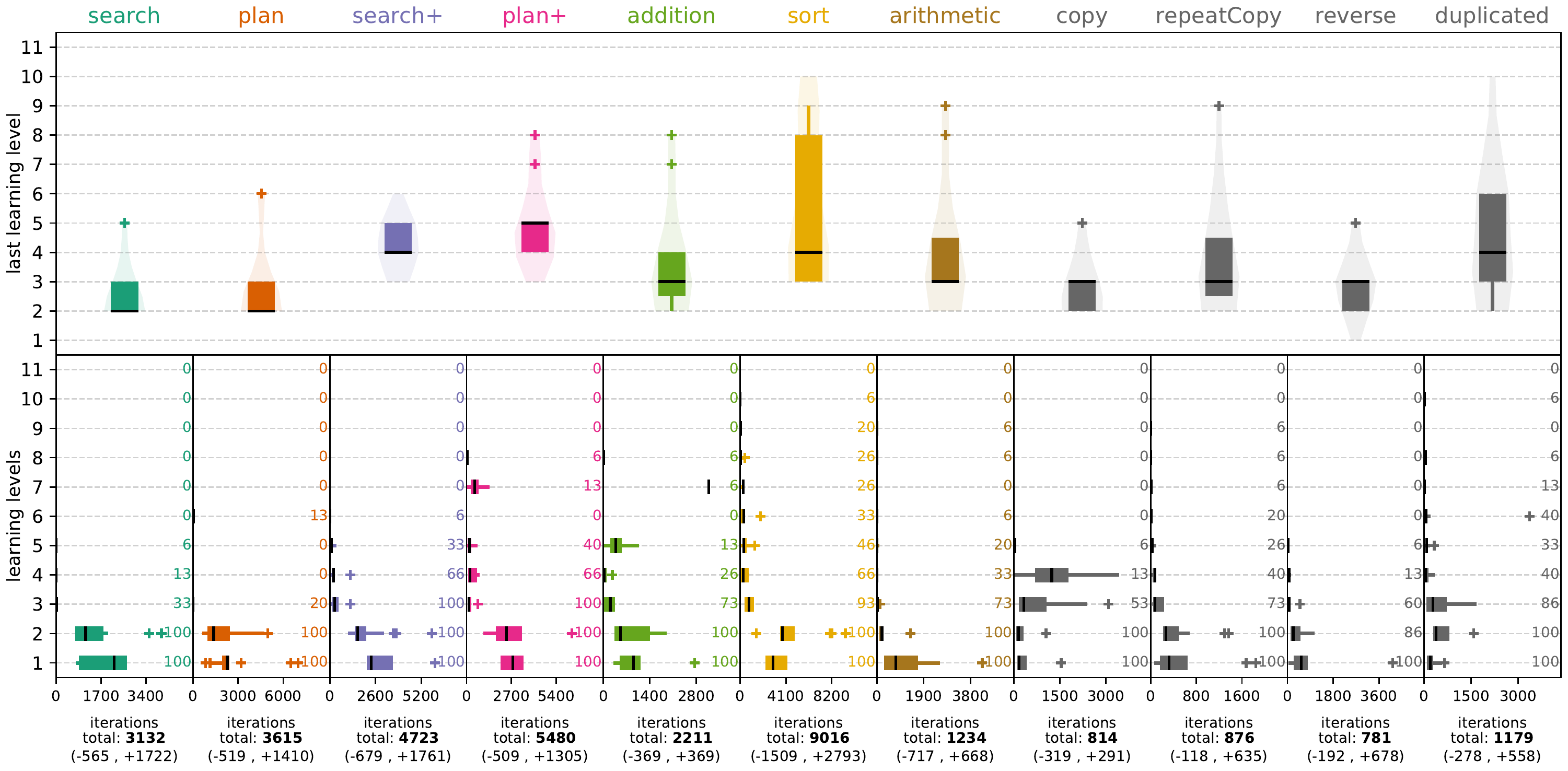}
	\caption{
	\textbf{Learning overview of all 11 learned algorithms.} 
	\textit{(Top)} The last curriculum level that triggered learning, i.e., where the last mistake occurred, is shown as the median (black line), the interquartile range (box) and outliers (plus).
	The shaded area shows the probability density of the data.
	All algorithms are learned within the first few levels and the solution generalizes to higher levels.
	\textit{(Bottom)} The number of training iterations per curriculum level is shown.
	The coloured numbers indicate the percentage of runs that triggered learning in each level.
	Learning occurs in the first levels, mostly within the first two, and subsequent levels only need a small amount of iterations to adapt, if at all.
	The total iterations show median and distances to the interquartile range of the total number of learning iterations.
	Results are obtained over $15$ runs for each algorithm.	
	} 
	\label{fig:learning} 
\end{figure*}
}

\newcommand{\figureTasks}{
\begin{figure*}[!t]
	\centering  
	\includegraphics[width=0.99\textwidth]{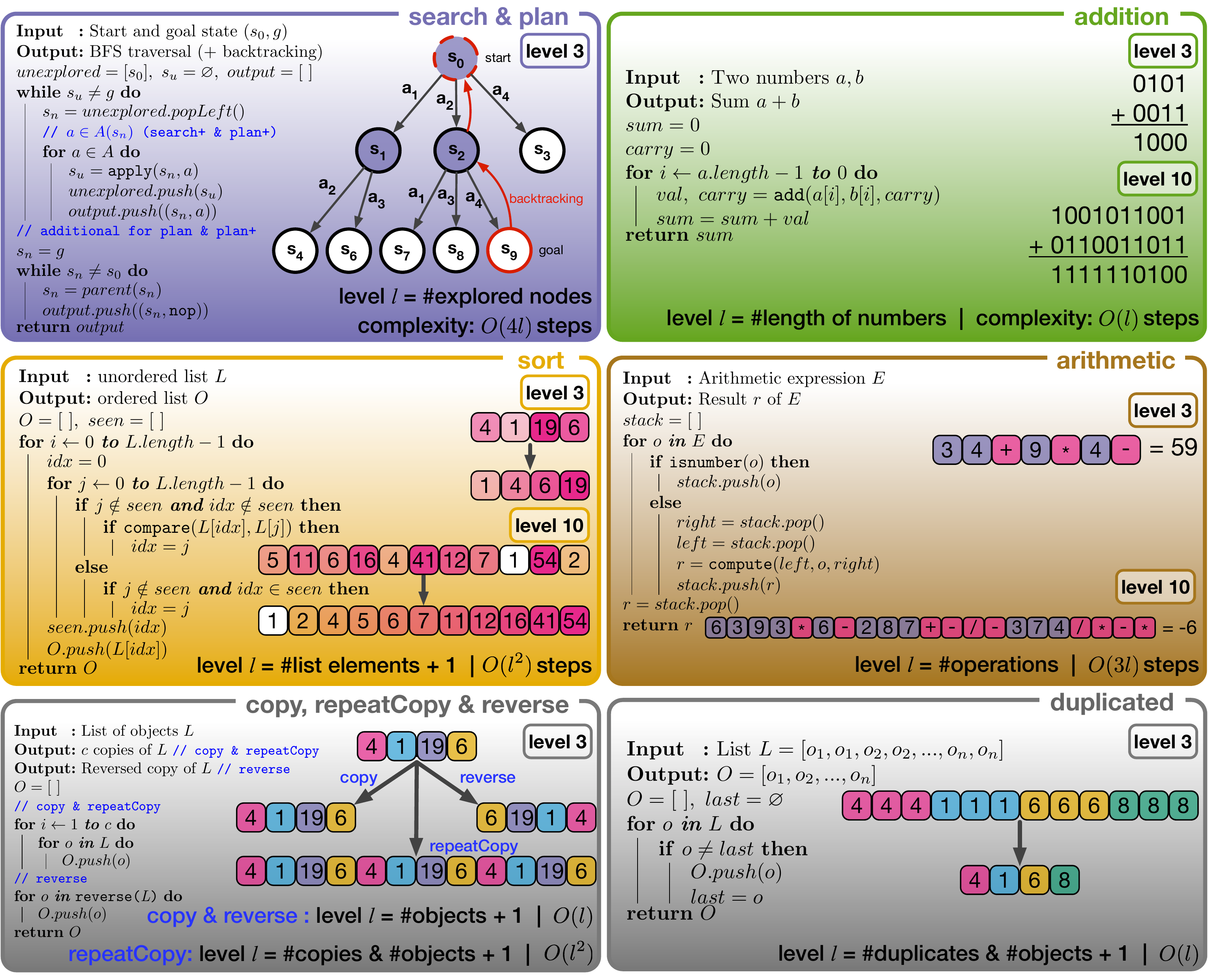}
	\caption{
	\textbf{Overview of the learned algorithms.} 
	All considered algorithms to learn are shown with their pseudocode, how their curriculum level complexity is defined, how the step complexity scales with the level, and examples from indicated levels.
	Note, the step complexity indicates the runtime complexity and only considers the steps after the input data is shown, neither taking the complexity of the data manipulation into account, nor the structure learning required while the input data is presented.
	} 
	\label{fig:tasks} 
\end{figure*}
}

\newcommand{\figureTransfers}{
\begin{figure*}[!t]
	\centering  
	\includegraphics[width=0.96\textwidth]{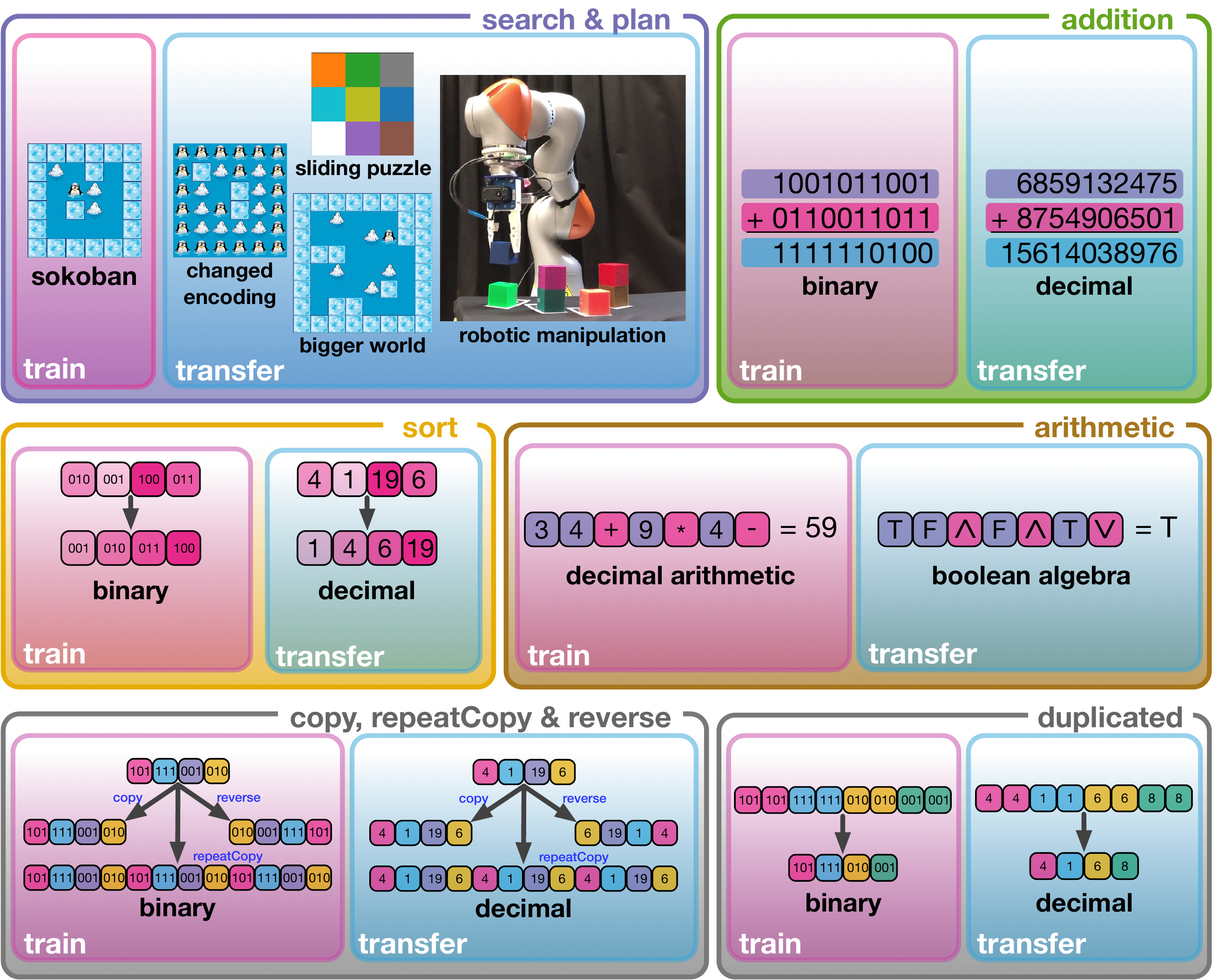}
	\caption{
	\textbf{Overview of the transfers of the learned algorithms.} 
	To show the abstract nature of the learned algorithms, each learned algorithm was transferred and tested on at least one different data representation or domain.
	All transfers were successful, i.e., the learned algorithm solved all samples in the new domain without triggering learning of the algorithmic modules, indicating the fulfilment of R2-R3.}
	 
	\label{fig:transfers} 
\end{figure*}
}

\newcommand{\figureBehavior}{
\begin{figure*}[!t]
	\centering  
	\includegraphics[width=0.99\textwidth]{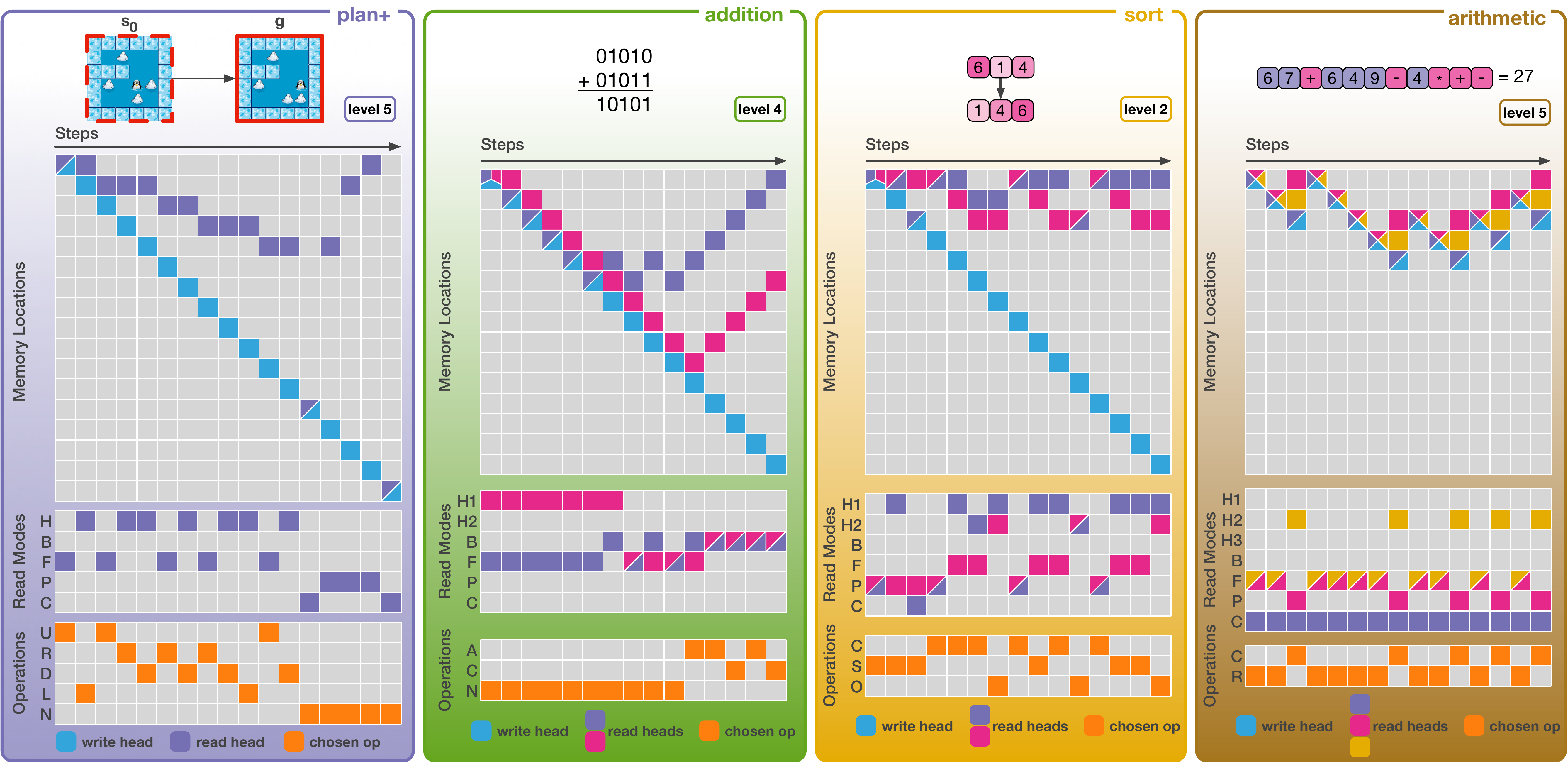}
	\caption{
	\textbf{Learned algorithmic behaviour of the NHC.} 
	The learned behaviour for four algorithms solving the examples shown at the top is illustrated.
	Shown are the written and read memory locations, the used read mode for each read head, and the operation signal sent to the ALU.
	First an example from the (\textit{plan+}) task. 
	The algorithm first builds the search tree by applying all applicable operations in a state and then shifts reading to the next state until the goal is found, then it backtracks the solution.
	Next an example from the (\textit{addition}) task.
	First, the two numbers to be added are presented after each other and are just stored.
	Then the two numbers are traversed from the low to the high end in parallel, adding the corresponding digits including possible carry bits.
	Next an example from the (\textit{sort}) task.
	After reading the unsorted list, the algorithm iterates over the list, finding and outputting the smallest element in each iteration.
	Lastly an example from the (\textit{arithmetic}) task.
	In the arithmetic task, the free gates were activated and the model learned to reuse memory locations in order to emulate the behaviour of a stack.
	The read heads always keep track of the head of the stack and when an arithmetic operation should be applied, it \textit{pops} the two top elements from the stack, which are then combined by the ALU according to the read operation.
	} 
	\label{fig:behavior} 
\end{figure*}
}

\newcommand{\tableLearning}{
\begin{table*}[!t]
\centering
\begin{adjustbox}{max width=\textwidth}
\begin{tabular}{rrr|c|c|c|c|c|c|c|c|c|c|c}
& & & search & plan & search+ & plan+ & addition & sort & arithmetic & copy & repeatCopy & reverse & duplicated \\
\cmidrule[2pt]{1-14}
\parbox[t]{3mm}{\multirow{6.5}{*}{\rotatebox[origin=c]{90}{{\large NHC}}}} 
& \parbox[t]{3mm}{\multirow{3}{*}{\rotatebox[origin=c]{90}{train}}} 
& iterations 
& 3132 ${\scriptstyle_{-565}^{+1722}}$ 
& 3615 ${\scriptstyle_{-519}^{+1410}}$ 
& 4723 ${\scriptstyle_{-679}^{+1761}}$ 
& 5480 ${\scriptstyle_{-509}^{+1305}}$ 
& 2211 ${\scriptstyle_{-369}^{+369}}$ 
& 9016 ${\scriptstyle_{-1509}^{+2793}}$ 
& 1234 ${\scriptstyle_{-717}^{+668}}$ 
& 814 ${\scriptstyle_{-319}^{+291}}$ 
& 876 ${\scriptstyle_{-118}^{+635}}$ 
& 781 ${\scriptstyle_{-192}^{+678}}$ 
& 1179 ${\scriptstyle_{-278}^{+558}}$ 
\\
\hhline{~~-|-|-|-|-|-|-|-|-|-|-|-}
& & last level 
& 2 ${\scriptstyle_{-0}^{+1}}$ ~~ 11 ${\scriptstyle_{-0}^{+0}}$
& 2 ${\scriptstyle_{-0}^{+1}}$ ~~ 11 ${\scriptstyle_{-0}^{+0}}$
& 4 ${\scriptstyle_{-0}^{+1}}$ ~~ 11 ${\scriptstyle_{-0}^{+0}}$
& 5 ${\scriptstyle_{-1}^{+0}}$ ~~ 11 ${\scriptstyle_{-0}^{+0}}$
& 3 ${\scriptstyle_{-0.5}^{+1}}$ ~~ 11 ${\scriptstyle_{-0}^{+0}}$
& 4 ${\scriptstyle_{-1}^{+4}}$ ~~ 11 ${\scriptstyle_{-0}^{+0}}$
& 3 ${\scriptstyle_{-0}^{+1.5}}$ ~~ 11 ${\scriptstyle_{-0}^{+0}}$
& 3 ${\scriptstyle_{-1}^{+0}}$ ~~ 11 ${\scriptstyle_{-0}^{+0}}$
& 3 ${\scriptstyle_{-0.5}^{+1.5}}$ ~~ 11 ${\scriptstyle_{-0}^{+0}}$
& 3 ${\scriptstyle_{-1}^{+0}}$ ~~ 11 ${\scriptstyle_{-0}^{+0}}$
& 4 ${\scriptstyle_{-1}^{+2}}$ ~~ 11 ${\scriptstyle_{-0}^{+0}}$
\\
\hhline{~~-|-|-|-|-|-|-|-|-|-|-|-}
& & level 10  \cellcolor{black!5}
& 100\% ~~ 100\% \cellcolor{black!5}
& 100\% ~~ 100\% \cellcolor{black!5}
& 100\% ~~ 100\% \cellcolor{black!5}
& 100\% ~~ 100\% \cellcolor{black!5}
& 100\% ~~ 100\% \cellcolor{black!5}
& 100\% ~~ 100\% \cellcolor{black!5}
& 100\% ~~ 100\%  \cellcolor{black!5}
& 100\% ~~ 100\%  \cellcolor{black!5}
& 100\% ~~ 100\%  \cellcolor{black!5}
& 100\% ~~ 100\% \cellcolor{black!5}
& 100\% ~~ 100\% \cellcolor{black!5}
\\
\cmidrule[1.5pt]{2-14}
& \parbox[t]{3mm}{\multirow{2.8}{*}{\rotatebox[origin=c]{90}{test}}}  
& level 100 \cellcolor{black!10}
& 100\% ~~ 100\% \cellcolor{black!10}
& 100\% ~~ 100\% \cellcolor{black!10}
& 100\% ~~ 100\% \cellcolor{black!10}
& 100\% ~~ 100\% \cellcolor{black!10}
& 80\% ~~ 80\% \cellcolor{black!10}
& 73.3\% ~~ 86.1\% \cellcolor{black!10}
& 100\% ~~ 100\% \cellcolor{black!10}
& 93.3\% ~~ 93.3\% \cellcolor{black!10}
& 80\% ~~ 80\% \cellcolor{black!10}
& 100\% ~~ 100\% \cellcolor{black!10}
& 100\% ~~ 100\% \cellcolor{black!10}
\\
\hhline{~~-|-|-|-|-|-|-|-|-|-|-|-}
& 
& level 500  \cellcolor{black!20}
& 100\% ~~ 100\% \cellcolor{black!20}
& 100\% ~~ 100\% \cellcolor{black!20}
& 100\% ~~ 100\% \cellcolor{black!20}
& 100\% ~~ 100\% \cellcolor{black!20}
& 80\% ~~ 80\%  \cellcolor{black!20}
& 73.3\% ~~ 87.7\% \cellcolor{black!20} 
& 100\% ~~ 100\% \cellcolor{black!20}
& 93.3\% ~~ 93.3\% \cellcolor{black!20}
& 80\% ~~ 80\% \cellcolor{black!20}
& 100\% ~~ 100\% \cellcolor{black!20}
& 100\% ~~ 100\% \cellcolor{black!20}
\\
\hhline{~~-|-|-|-|-|-|-|-|-|-|-|-}
&  
& level 1000  \cellcolor{black!30}
& 100\% ~~ 100\% \cellcolor{black!30}
& 100\% ~~ 100\% \cellcolor{black!30}
& 100\% ~~ 100\% \cellcolor{black!30}
& 100\% ~~ 100\% \cellcolor{black!30}
& 80\% ~~ 80\% \cellcolor{black!30}
& 73.3\% ~~ 86.1\%  \cellcolor{black!30}
& 100\% ~~ 100\% \cellcolor{black!30}
& 93.3\% ~~ 93.3\% \cellcolor{black!30}
& 80\% ~~ 80\% \cellcolor{black!30}
& 100\% ~~ 100\% \cellcolor{black!30}
& 100\% ~~ 100\% \cellcolor{black!30}
\\
\cmidrule[2pt]{1-14}
\parbox[t]{3mm}{\multirow{6.5}{*}{\rotatebox[origin=c]{90}{{\large NHC-anc}}}} 
& \parbox[t]{3mm}{\multirow{3}{*}{\rotatebox[origin=c]{90}{train}}} 
& iterations 
& 4319 ${\scriptstyle_{-888}^{+448}}$
& --
& -- 
& --
& 1586 ${\scriptstyle_{-488}^{+1410}}$
& 5677 ${\scriptstyle_{-1311}^{+892}}$  
& 2467 ${\scriptstyle_{-1041}^{+494}}$  
& 1051 ${\scriptstyle_{-159}^{+841}}$ 
& 3496 ${\scriptstyle_{-697}^{+648}}$ 
& 848 ${\scriptstyle_{-558}^{+717}}$ 
& 2465 ${\scriptstyle_{-132}^{+607}}$ 
\\
\hhline{~~-|-|-|-|-|-|-|-|-|-|-|-}
& & last level 
& 1 ${\scriptstyle_{-0}^{+0}}$ ~~  0 ${\scriptstyle_{-0}^{+0}}$ 
& --
& --
& --
& 3 ${\scriptstyle_{-1}^{+1}}$ ~~ 11 ${\scriptstyle_{-0}^{+0}}$
& 2 ${\scriptstyle_{-1}^{+0}}$ ~~  1 ${\scriptstyle_{-1}^{+0}}$
& 3 ${\scriptstyle_{-1}^{+1.5}}$ ~~ 11 ${\scriptstyle_{-0}^{+0}}$
& 2 ${\scriptstyle_{-0}^{+0}}$	~~ 11 ${\scriptstyle_{-0}^{+0}}$
& 2 ${\scriptstyle_{-0}^{+0}}$ ~~ 1 ${\scriptstyle_{-1}^{+0}}$
& 3 ${\scriptstyle_{-1}^{+1}}$ ~~ 11 ${\scriptstyle_{-0}^{+0}}$
& 2 ${\scriptstyle_{-0}^{+0}}$ ~~  1 ${\scriptstyle_{-0}^{+0}}$
\\
\hhline{~~-|-|-|-|-|-|-|-|-|-|-|-}
& & level 10 \cellcolor{black!5}
& 0\% ~~ 0\% \cellcolor{black!5}
& --  \cellcolor{black!5}
& --  \cellcolor{black!5}
& -- \cellcolor{black!5}
& 100\% ~~ 100\% \cellcolor{black!5}
& 0\% ~~ 0\% \cellcolor{black!5}
& 86.7\% ~~ 86.7\% \cellcolor{black!5}
& 86.7\% ~~ 86.7\% \cellcolor{black!5}
& 0\% ~~ 1.9\% \cellcolor{black!5}
& 100\% ~~ 100\% \cellcolor{black!5}
& 0\% ~~ 2.86\% \cellcolor{black!5}
\\
\cmidrule[1.5pt]{2-14}
& \parbox[t]{3mm}{\multirow{2.8}{*}{\rotatebox[origin=c]{90}{test}}}  
& level 100  \cellcolor{black!10}
& -- \cellcolor{black!10}
& -- \cellcolor{black!10}
& -- \cellcolor{black!10}
& -- \cellcolor{black!10}
& 93.9\% ~~ 93.3\% \cellcolor{black!10}
& -- \cellcolor{black!10}
& 80\% ~~ 86.3\% \cellcolor{black!10}
& 86.7\% ~~ 86.7\% \cellcolor{black!10}
& 0\% ~~ 0\% \cellcolor{black!10}
& 93.3\% ~~ 93.3\% \cellcolor{black!10}
& 0\% ~~ 0\% \cellcolor{black!10}
\\
\hhline{~~-|-|-|-|-|-|-|-|-|-|-|-}
& & level 500  \cellcolor{black!20}
& -- \cellcolor{black!20}
& -- \cellcolor{black!20}
& -- \cellcolor{black!20}
& -- \cellcolor{black!20}
& 93.9\% ~~ 93.3\% \cellcolor{black!20}
& -- \cellcolor{black!20}
& 80\% ~~ 82.3\% \cellcolor{black!20}
& 86.7\% ~~ 86.7\% \cellcolor{black!20}
& -- \cellcolor{black!20}
& 93.3\% ~~ 93.3\% \cellcolor{black!20}
& -- \cellcolor{black!20}
\\
\hhline{~~-|-|-|-|-|-|-|-|-|-|-|-}
& & level 1000  \cellcolor{black!30}
& -- \cellcolor{black!30}
& -- \cellcolor{black!30}
& -- \cellcolor{black!30}
& -- \cellcolor{black!30}
& 93.9\% ~~ 93.3\% \cellcolor{black!30}
& -- \cellcolor{black!30}
& 80\% ~~ 80.1\% \cellcolor{black!30}
& 86.7\% ~~ 86.7\% \cellcolor{black!30}
& -- \cellcolor{black!30}
& 93.3\% ~~ 93.3\% \cellcolor{black!30}
& -- \cellcolor{black!30}
\\
\cmidrule[2pt]{1-14}
\parbox[t]{3mm}{\multirow{6.5}{*}{\rotatebox[origin=c]{90}{{\large NHC-prev}}}} 
& \parbox[t]{3mm}{\multirow{3}{*}{\rotatebox[origin=c]{90}{train}}} 
& iterations 
& 3913 ${\scriptstyle_{-749}^{+937}}$
& --
& -- 
& --
& 2722 ${\scriptstyle_{-868}^{+765}}$
& 3784 ${\scriptstyle_{-564}^{+1002}}$  
& 2391 ${\scriptstyle_{-824}^{+1432}}$  
& 362 ${\scriptstyle_{-123}^{+163}}$ 
& 1118 ${\scriptstyle_{-600}^{+267}}$ 
& 322 ${\scriptstyle_{-118}^{+762}}$ 
& 2398 ${\scriptstyle_{-275}^{+235}}$ 
\\
\hhline{~~-|-|-|-|-|-|-|-|-|-|-|-}
& & last level 
& 1 ${\scriptstyle_{-0}^{+0}}$ ~~  0 ${\scriptstyle_{-0}^{+0}}$ 
& --
& --
& --
& 3 ${\scriptstyle_{-1}^{+0}}$ ~~ 11 ${\scriptstyle_{-0}^{+0}}$
& 1 ${\scriptstyle_{-0}^{+0}}$ ~~  0 ${\scriptstyle_{-0}^{+0}}$
& 6 ${\scriptstyle_{-0.5}^{+0}}$ ~~  11 ${\scriptstyle_{-0}^{+0}}$
& 2 ${\scriptstyle_{-0}^{+1.5}}$	~~ 11 ${\scriptstyle_{-0}^{+0}}$
& 3 ${\scriptstyle_{-1}^{+0}}$ ~~ 11 ${\scriptstyle_{-0}^{+0}}$
& 3 ${\scriptstyle_{-1}^{+0.5}}$ ~~ 11 ${\scriptstyle_{-0}^{+0}}$
& 2 ${\scriptstyle_{-0.5}^{+0}}$ ~~  1 ${\scriptstyle_{-0.5}^{+0}}$
\\
\hhline{~~-|-|-|-|-|-|-|-|-|-|-|-}
& & level 10  \cellcolor{black!5}
& 0\% ~~ 0\% \cellcolor{black!5}
& --  \cellcolor{black!5}
& --  \cellcolor{black!5}
& -- \cellcolor{black!5}
& 100\% ~~ 100\% \cellcolor{black!5}
& 0\% ~~ 0\% \cellcolor{black!5}
& 93.3\% ~~ 93.3\% \cellcolor{black!5}
& 100\% ~~ 100\% \cellcolor{black!5}
& 100\% ~~ 100\% \cellcolor{black!5}
& 86.7\% ~~ 86.7\% \cellcolor{black!5}
& 0\% ~~ 0.7\% \cellcolor{black!5}
\\
\cmidrule[1.5pt]{2-14}
& \parbox[t]{3mm}{\multirow{2.8}{*}{\rotatebox[origin=c]{90}{test}}}  
& level 100  \cellcolor{black!10}
& -- \cellcolor{black!10}
& -- \cellcolor{black!10}
& -- \cellcolor{black!10}
& -- \cellcolor{black!10}
& 100\% ~~ 100\% \cellcolor{black!10}
& -- \cellcolor{black!10}
& 93.3\% ~~ 93.3\% \cellcolor{black!10}
& 100\% ~~ 100\% \cellcolor{black!10}
& 93.3\% ~~ 93.3\% \cellcolor{black!10}
& 86.7\% ~~ 86.7\% \cellcolor{black!10}
& 0\% ~~ 0\% \cellcolor{black!10}
\\
\hhline{~~-|-|-|-|-|-|-|-|-|-|-|-}
& & level 500  \cellcolor{black!20}
& -- \cellcolor{black!20}
& -- \cellcolor{black!20}
& -- \cellcolor{black!20}
& -- \cellcolor{black!20}
& 100\% ~~ 100\% \cellcolor{black!20}
& -- \cellcolor{black!20}
& 86.7\% ~~ 93.2\% \cellcolor{black!20}
& 100\% ~~ 100\% \cellcolor{black!20}
& 93.3\% ~~ 93.3\% \cellcolor{black!20}
& 86.7\% ~~ 86.7\% \cellcolor{black!20}
& -- \cellcolor{black!20}
\\
\hhline{~~-|-|-|-|-|-|-|-|-|-|-|-}
& & level 1000  \cellcolor{black!30}
& -- \cellcolor{black!30}
& -- \cellcolor{black!30}
& -- \cellcolor{black!30}
& -- \cellcolor{black!30}
& 100\% ~~ 100\% \cellcolor{black!30}
& -- \cellcolor{black!30}
& 86.7\% ~~ 93.2\% \cellcolor{black!30}
& 100\% ~~ 100\% \cellcolor{black!30}
& 93.3\% ~~ 93.3\% \cellcolor{black!30}
& 86.7\% ~~ 86.7\% \cellcolor{black!30}
& -- \cellcolor{black!30}
\\
\cmidrule[2pt]{1-14}
\parbox[t]{3mm}{\multirow{6.5}{*}{\rotatebox[origin=c]{90}{{\large DNC+is+ha}}}} 
& \parbox[t]{3mm}{\multirow{3}{*}{\rotatebox[origin=c]{90}{train}}} 
& iterations 
& 3159 ${\scriptstyle_{-749}^{+937}}$
& --
& -- 
& --
& 4315 ${\scriptstyle_{-365}^{+1584}}$
& 5539 ${\scriptstyle_{-905}^{+2653}}$  
& 4360 ${\scriptstyle_{-845}^{+1842}}$  
& 4355 ${\scriptstyle_{-1121}^{+945}}$ 
& 4163 ${\scriptstyle_{-999}^{+475}}$ 
& 2126 ${\scriptstyle_{-123}^{+766}}$ 
& 3756 ${\scriptstyle_{-1046}^{+2025}}$ 
\\
\hhline{~~-|-|-|-|-|-|-|-|-|-|-|-}
& & last level 
& 1 ${\scriptstyle_{-0}^{+0}}$ ~~  0 ${\scriptstyle_{-0}^{+0}}$ 
& --
& --
& --
& 10 ${\scriptstyle_{-0}^{+0}}$ ~~ 11 ${\scriptstyle_{-0}^{+0}}$
&  1 ${\scriptstyle_{-0}^{+0}}$ ~~  0 ${\scriptstyle_{-0}^{+0}}$
&  2 ${\scriptstyle_{-0}^{+1}}$ ~~  1 ${\scriptstyle_{-1}^{+0}}$
& 10 ${\scriptstyle_{-1}^{+0}}$	~~ 11 ${\scriptstyle_{-0}^{+0}}$
& 10 ${\scriptstyle_{-0}^{+0}}$ ~~ 11 ${\scriptstyle_{-0}^{+0}}$
& 10 ${\scriptstyle_{-1}^{+0}}$ ~~ 11 ${\scriptstyle_{-0}^{+0}}$
&  2 ${\scriptstyle_{-0}^{+1}}$ ~~  1 ${\scriptstyle_{-1}^{+0}}$
\\
\hhline{~~-|-|-|-|-|-|-|-|-|-|-|-}
& & level 10  \cellcolor{black!5}
& 0\% ~~ 0\% \cellcolor{black!5}
& --  \cellcolor{black!5}
& --  \cellcolor{black!5}
& -- \cellcolor{black!5}
& 20\% ~~ 65.2\% \cellcolor{black!5}
& 0\% ~~ 0\% \cellcolor{black!5}
& 0\% ~~ 0\% \cellcolor{black!5}
& 100\% ~~ 100\% \cellcolor{black!5}
& 20\% ~~ 92\% \cellcolor{black!5}
& 100\% ~~ 100\% \cellcolor{black!5}
& 0\% ~~ 5.7\% \cellcolor{black!5}
\\
\cmidrule[1.5pt]{2-14}
& \parbox[t]{3mm}{\multirow{2.8}{*}{\rotatebox[origin=c]{90}{test}}}  
& level 100  \cellcolor{black!10}
& -- \cellcolor{black!10}
& -- \cellcolor{black!10}
& -- \cellcolor{black!10}
& -- \cellcolor{black!10}
& 0\% ~~ 0\% \cellcolor{black!10}
& -- \cellcolor{black!10}
& -- \cellcolor{black!10}
& 0\% ~~ 0\% \cellcolor{black!10}
& 13.3\% ~~ 13.3\% \cellcolor{black!10}
& 0\% ~~ 0\% \cellcolor{black!10}
& 0\% ~~ 0\% \cellcolor{black!10}
\\
\hhline{~~-|-|-|-|-|-|-|-|-|-|-|-}
& & level 500  \cellcolor{black!20}
& -- \cellcolor{black!20}
& -- \cellcolor{black!20}
& -- \cellcolor{black!20}
& -- \cellcolor{black!20}
& -- \cellcolor{black!20}
& -- \cellcolor{black!20}
& -- \cellcolor{black!20}
& -- \cellcolor{black!20}
& 13.3\% ~~ 13.3\% \cellcolor{black!20}
& -- \cellcolor{black!20}
& -- \cellcolor{black!20}
\\
\hhline{~~-|-|-|-|-|-|-|-|-|-|-|-}
& & level 1000  \cellcolor{black!30}
& -- \cellcolor{black!30}
& -- \cellcolor{black!30}
& -- \cellcolor{black!30}
& -- \cellcolor{black!30}
& -- \cellcolor{black!30}
& -- \cellcolor{black!30}
& -- \cellcolor{black!30}
& -- \cellcolor{black!30}
& 13.3\% ~~ 13.3\% \cellcolor{black!30}
& -- \cellcolor{black!30}
& -- \cellcolor{black!30}
\\
\cmidrule[2pt]{1-14}
\parbox[t]{3mm}{\multirow{6.5}{*}{\rotatebox[origin=c]{90}{{\Large DNC}\supercite{graves2016hybrid}}}} 
& \parbox[t]{3mm}{\multirow{3}{*}{\rotatebox[origin=c]{90}{train}}} 
& iterations 
& 500k
& --
& --
& --
& 500k
& 500k
& 500k
& 500k
& 500k
& 500k
& 500k
\\
\hhline{~~-|-|-|-|-|-|-|-|-|-|-|-}
& & last level 
& 1 ${\scriptstyle_{-0}^{+0}}$ ~~  0 ${\scriptstyle_{-0}^{+0}}$ 
& --
& --
& --
& 6 ${\scriptstyle_{-0}^{+0}}$ ~~  5 ${\scriptstyle_{-0}^{+0}}$ 
& 2 ${\scriptstyle_{-0}^{+0}}$ ~~  1 ${\scriptstyle_{-0}^{+0}}$
& 2 ${\scriptstyle_{-0}^{+0}}$ ~~  1 ${\scriptstyle_{-0}^{+0}}$
& 6 ${\scriptstyle_{-0}^{+0}}$ ~~  5 ${\scriptstyle_{-0}^{+0}}$ 
& 2 ${\scriptstyle_{-0}^{+0}}$ ~~  1 ${\scriptstyle_{-0}^{+0}}$ 
& 6 ${\scriptstyle_{-0}^{+0}}$ ~~  5 ${\scriptstyle_{-0}^{+0}}$ 
& 5 ${\scriptstyle_{-0}^{+0}}$ ~~  4 ${\scriptstyle_{-0}^{+0}}$ 
\\
\hhline{~~-|-|-|-|-|-|-|-|-|-|-|-}
& & level 10  \cellcolor{black!5}
& 0\% ~~  0\% \cellcolor{black!5}
& -- \cellcolor{black!5}
& -- \cellcolor{black!5}
& -- \cellcolor{black!5}
& 0\% ~~  0\% \cellcolor{black!5}
& 0\% ~~  0\% \cellcolor{black!5}
& 0\% ~~  0\% \cellcolor{black!5}
& 0\% ~~  0\% \cellcolor{black!5}
& 0\% ~~  0\% \cellcolor{black!5}
& 13.3\% ~~ 13.3\% \cellcolor{black!5}
& 0\% ~~  0\% \cellcolor{black!5}
\\
\cmidrule[1.5pt]{2-14}
& \parbox[t]{3mm}{\multirow{2.8}{*}{\rotatebox[origin=c]{90}{test}}}  
& level 100  \cellcolor{black!10}
& -- \cellcolor{black!10}
& -- \cellcolor{black!10}
& -- \cellcolor{black!10}
& -- \cellcolor{black!10}
& -- \cellcolor{black!10}
& -- \cellcolor{black!10}
& -- \cellcolor{black!10}
& -- \cellcolor{black!10}
& -- \cellcolor{black!10}
& 0\% ~~  0\% \cellcolor{black!10}
& -- \cellcolor{black!10}
\\
\hhline{~~-|-|-|-|-|-|-|-|-|-|-|-}
& & level 500  \cellcolor{black!20}
& -- \cellcolor{black!20}
& -- \cellcolor{black!20}
& -- \cellcolor{black!20}
& -- \cellcolor{black!20}
& -- \cellcolor{black!20}
& -- \cellcolor{black!20}
& -- \cellcolor{black!20}
& -- \cellcolor{black!20}
& -- \cellcolor{black!20}
& -- \cellcolor{black!20}
& --  \cellcolor{black!20}
\\
\hhline{~~-|-|-|-|-|-|-|-|-|-|-|-}
& & level 1000  \cellcolor{black!30}
& -- \cellcolor{black!30}
& -- \cellcolor{black!30}
& -- \cellcolor{black!30}
& -- \cellcolor{black!30}
& -- \cellcolor{black!30}
& -- \cellcolor{black!30}
& -- \cellcolor{black!30}
& -- \cellcolor{black!30}
& -- \cellcolor{black!30}
& -- \cellcolor{black!30}
& --  \cellcolor{black!30}
\\
\cmidrule[2pt]{1-14}
\end{tabular}
\end{adjustbox}
\caption{\textbf{Evaluation and Comparison.}
Shown are results over $15$ runs for each algorithm and model.
Triplets like 5 ${\scriptstyle_{-2}^{+3}}$ show median and distances to the interquartile range.
\textit{Iterations} refers to the number of learning iterations.
The two \textit{last level} triplets show the last learning level (left) and the last solved level (right).
The two percentages in \textit{level X} indicate the amount of perfect runs (left), i.e., runs that solved all presented samples, and the amount of solved samples over all runs (right).
All NHC variants and the DNC+is+ha model are using the information split and data modules, and are trained with NES.
The original DNC is trained in a supervised setting with backpropagation.
}
\label{table:comparison}
\end{table*} 
}

\newcommand{\figureCurves}{
\begin{figure*}[!t]
\centering  
\includegraphics[width=0.99\textwidth]{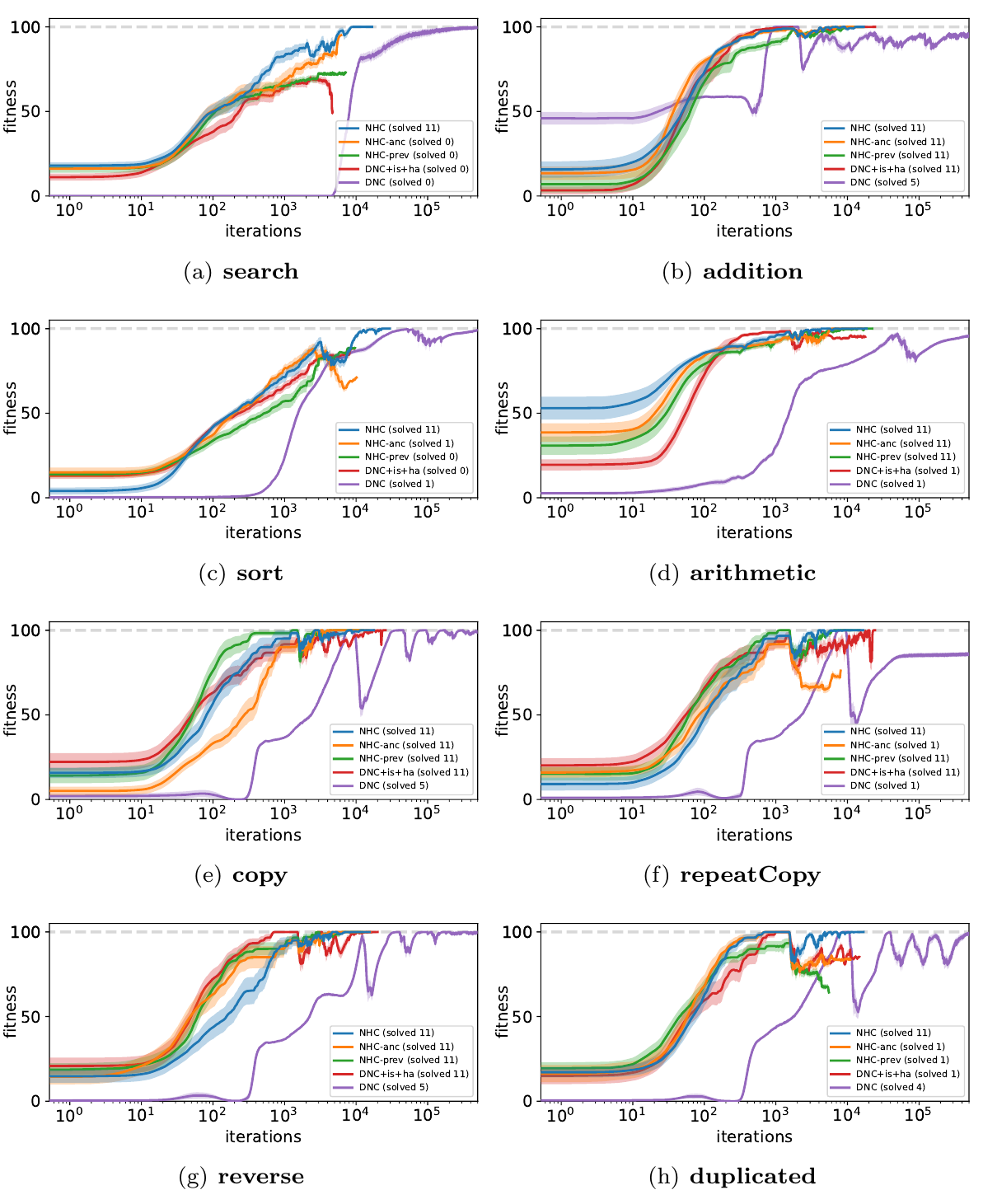}
\vspace{-2pt}
\caption{
\textbf{Learning curves comparison.} Shown are the mean and the standard error of the fitness during learning over $15$ runs.
Note the log-scale of the x-axis.
Solved \textit{X} in the legend indicates the median solved level.
The full NHC is the only model that successfully learns all algorithms reliably.
More details on these evaluations are given in Table~\ref*{table:comparison}.
} 
\label{fig:curves} 
\end{figure*}
}

\begin{document}

\begin{abstract}
A key feature of intelligent behaviour is the ability to learn abstract strategies that scale and transfer to unfamiliar problems.
An abstract strategy solves every sample from a problem class, no matter its representation or complexity -- like algorithms in computer science.
Neural networks are powerful models for processing sensory data, discovering hidden patterns, and learning complex functions, but they struggle to learn such iterative, sequential or hierarchical algorithmic strategies.
Extending neural networks with external memories has increased their capacities in learning such strategies, but they are still prone to data variations, struggle to learn scalable and transferable solutions, and require massive training data.
We present the Neural Harvard Computer (NHC), a memory-augmented network based architecture, that employs abstraction by decoupling algorithmic operations from data manipulations, realized by splitting the information flow and separated modules.
This abstraction mechanism and evolutionary training enable the learning of robust and scalable algorithmic solutions.
On a diverse set of $11$ algorithms with varying complexities, we show that the NHC reliably learns algorithmic solutions with strong generalization and abstraction: perfect generalization and scaling to arbitrary task configurations and complexities far beyond seen during training, and being independent of the data representation and the task domain.
\end{abstract}

\flushbottom
\maketitle
\thispagestyle{firstpage}

\begin{refsegment}
\section*{Introduction}
\label{sec:intro} 
A crucial ability for intelligent behaviour is to transfer strategies from one problem to another, studied, for example, in the fields of lifelong and transfer learning~\supercite{taylor2009transfer,silver2013lifelong,weiss2016survey,parisi2019continual}.
Learning and especially deep learning systems have been shown to learn a variety of complex specialized tasks~\supercite{schmidhuber2015deep,mnih2015human,silver2016mastering,fawaz2019deep,botvinick2019reinforcement,liu2020deep}, but extracting the underlying structure of the solution for effective transfer is an open research question~\supercite{weiss2016survey}.

The key for effective transfer, and a main pillar of (human) intelligence, is the concept of structure and abstraction~\supercite{tenenbaum2011grow,lake2017building,konidaris2019necessity}.
To study the learning of such abstract strategies, the concept of \textit{algorithms} like in computer science~\supercite{cormen2009introduction} is an ideal example for such transferable, abstract and structured solution strategies.
 
An algorithm is a sequence of instructions, which often represent solutions to smaller subproblems.
This sequence of instructions solves a given problem when executed, independent of the specific instantiation of the problem.
For example, consider the task of sorting a set of objects. 
The algorithmic solution, specified as the sequence of instructions, is able to sort any number of arbitrary classes of objects in any order, e.g., toys by colour, waste by type, or numbers by value, by using \textit{the same sequence of instructions}, as long as the features and comparison operators defining the order are specified.

Learning such structured, abstract strategies enables the effective transfer to new domains and representations as the abstract solution is independent of both.
In contrast, transfer learning usually focuses on improving learning speed on a new task by leveraging knowledge from previously learned tasks, whereas algorithmic solutions do not need to (re)learn at all, only the data specific operations need to adapt. 
In other words, the sequence of instructions does not need to be adapted, only the instructions, i.e., the solutions to smaller subproblems.
Moreover, such structured abstract strategies have built-in generalization capabilities to new task configurations and complexities, and can be interpreted better than, for example, common blackbox models like deep end-to-end networks.

\subsection*{The Problem of Learning Algorithmic Solutions}
To study the learning of such abstract and structured strategies, we investigate the problem of learning algorithmic solutions which we characterize by three requirements:
\begin{itemize}
\item \textbf{R1} -- generalization and scaling to different and unseen task configurations and complexities
\item \textbf{R2} -- independence of the data representation
\item \textbf{R3} -- independence of the task domain
\end{itemize} 
 
Picking up the sorting algorithm example again, R1 represents the generalization and scaling properties, which allow to sort lists of arbitrary length and initial order, while R2 and R3 represent the abstract nature of the solution.
This abstraction enables the algorithm, for example, to sort a list of binary numbers while being trained only on hexadecimal numbers (R2). 
Furthermore, the algorithm trained on numbers is able to sort lists of strings (R3).
If R1 -- R3 are fulfilled, the algorithmic solution does not need to be retrained or adapted to solve unforeseen task instantiations -- only the data specific operations need to be adjusted.

\figureNHC

Earlier research on solving algorithmic problems was done, for example, in grammar learning~\supercite{das1992learning,mozer1993connectionist,zeng1994discrete}, and is becoming a more and more active field in recent years outside of it~\supercite{graves2014neural,joulin2015inferring,graves2016hybrid,neelakantan2016neural,kaiser2016neural,zaremba2016learning,greve2016evolving,trask2018neural,Madsen2020Neural,Le2020Neural,reed2015neural,kurach2016neural,cai2017making,dong2019neural,velickovic2020neural}, with a typical focus on identifying algorithmic generated patterns or solving \textit{algorithmic problems} in an end-to-end setup~\supercite{graves2014neural,joulin2015inferring,graves2016hybrid,neelakantan2016neural,kaiser2016neural,zaremba2016learning,greve2016evolving,trask2018neural,Madsen2020Neural,Le2020Neural}, and less on finding \textit{algorithmic solutions}~\supercite{reed2015neural,kurach2016neural,cai2017making,dong2019neural,velickovic2020neural} that consider the three discussed requirements R1 -- R3 for generalization, scaling and abstraction.

While R1 is typically tackled in some (relaxed) form, as it represents the overall goal of generalization in machine learning, the abstraction abilities R2 and R3 are missing. 
Additionally, most algorithms require a form of feedback, using computed intermediate results from one computational step in subsequent steps, and a variable number of computational steps to solve a problem instance.
Thus, it is necessary to be able to cope with varying numbers of steps and determining when to stop, in contrast to using a fixed number of steps~\supercite{sukhbaatar2015end,neelakantan2016neural}, and to be able to re-use intermediate results, i.e., feeding back the models output as its input.
These features make the learning problem even more challenging.

\figureTasks
\section*{The Neural Harvard Computer}
\label{sec:intro_nhc}
The proposed Neural Harvard Computer (NHC) is a modular architecture that is based on memory-augmented neural networks~\supercite{das1992learning,mozer1993connectionist,zeng1994discrete,graves2014neural,joulin2015inferring,sukhbaatar2015end,weston2014memory,grefenstette2015learning,graves2016hybrid,neelakantan2016neural,zaremba2016learning,kumar2016ask,greve2016evolving,reed2015neural,kurach2016neural,wayne2018unsupervised,merrild2018hyperntm,khadka2019neuroevolution,Le2020Neural} and inspired by modern computer architectures (see Figure~\ref{fig:nhc} for a sketch of the NHC).
Memory-augmented networks add an external memory to a neural network, that allows to separate computation and memorization -- in classical neural networks both is encoded in the synaptic weights.

The external memory can be realized differently, e.g., as a memory matrix~\supercite{graves2016hybrid}, tape~\supercite{zaremba2016learning}, stack~\supercite{joulin2015inferring}, and the so called controller network can write and read information through a defined interface, that controls the memory access, e.g., moving the head one step to the right for a tape memory, or pop the top information in a stack memory.
In the NHC the external memory is realized as a matrix and interaction with the memory is done via write and read heads, similar to the Differential Neural Computer~\supercite{graves2016hybrid}.
These heads interact with the memory by writing or reading information into or from the memory matrix, where each row corresponds to a memory location with a specified word size, i.e., length of the information vector.

\paragraph{Information Split}
Learning algorithmic solutions requires the decoupling of algorithmic computations from data dependent manipulations and domain.
Therefore, an abstraction level is introduced by dividing the information flow into two streams, data $d$ and control stream $c$.
Like the introduction of external memories to neural networks helps to separate computation and memorization, the information flow split helps to separate algorithmic computations and data specific manipulations.

This information split induces two major features of the NHC: (1) the split into data modules that operate on the data stream $d$ and algorithmic modules that operate on the control stream $c$, and (2) the introduction of two coupled memories.
The algorithmic modules operate on the control stream $c$, i.e., $AlgModule(c) \longrightarrow c$, whereas the data modules \textit{Input} and \textit{ALU} are operating on the data stream $d$, i.e., $DataModule(d,c) \longrightarrow d,c$.
They create an abstract interface and \textit{separation} between algorithmic computation and data specific processing.
While the \textit{Input} module receives the external data and provides algorithm specific control signals, the \textit{ALU} receives data and control information to manipulate the data to create new data -- hence the name \textit{arithmetic logic unit} -- that is fed back to \textit{Input} to be available in the next computation step.
These two modules are data specific and need to be adapted for a new data representation or domain.

\paragraph{Algorithmic Modules}
The algorithmic modules consists of the \textit{Controller}, \textit{Memory} and \textit{Bus}.
These modules form the core of the NHC (see Figure~\ref{fig:nhc}) and are responsible for encoding and learning the algorithmic solution based on the control stream $c$.

\paragraph{The Controller} receives the control signals from the \textit{Input} and the information read from the memory in the previous step -- depending on the task to learn, it can also receive feedback from the \textit{Bus} and \textit{ALU}.
It learns an internal representation of the algorithm state that is send to the \textit{Memory} and \textit{Bus} modules.

\paragraph{The Memory} module uses this representation in addition to the control signals from the \textit{Input} to learn a set of interfaces for interacting with the memory matrices.
These learned interfaces control the write and read heads and hence, what information is accessed.
First the locations read in the previous step are potentially updated (prev), then new information is written via the write heads (write), and finally the read heads read information (read) that is send back to controller and the \textit{Bus}.
Write and read heads are using hard decisions, i.e., each head interacts with one memory location.

The abstraction introduced by the information split also creates the necessity to store data and control information separately.
Therefore, the \textit{Memory} module uses two memory matrices $M^c : N \times C$ and $M^d : N \times D$ to store the control and data information respectively, with $N$ locations, and $C$ and $D$ the word size accordingly.
The two memories are coupled such that in each step the same locations are accessed.
This allows to store algorithmic control information alongside the data information.
New information is written via the write heads to unused memory locations, and locations can be freed by free gates to be reused.
For reading information from the memory, there are several read modes to steer the read heads.

The \textit{HALT} modes move the read head to the previously read location of the associated head.
For example, with two read heads, each head can use the modes \textit{H1} and \textit{H2} to move to the location previously read by the corresponding head.

To read data in the order of which it was written, the DNC introduced a \textit{temporal linkage} mechanism that keeps track of the order of written locations.
The NHC uses a simplified version of this temporal linkage.
This temporal linkage provides two read modes, one to move the read head \textit{forward} to the location that was written next (F), and one to move the head \textit{backwards} to the location that was written before (B).

Algorithms often require hierarchical data structures or dependencies.
To provide such dependencies, the NHC employs an \textit{ancestry linkage} mechanism.
This mechanism keeps track for each written location, which location was read before.
Therefore, this mechanisms provides two read modes, one to move the read head to the \textit{parent} (P), the location that was read before, and one to move the read head to the \textit{child} (C), the location that was written.

\paragraph{The Bus} combines the representation learned by the \textit{Controller} with the information read out from the \textit{Memory} to produce the control signal that is send to the \textit{ALU}, indicating which operation to apply on the read data.
By using the information read from memory, the \textit{Bus} can incorporate this new information in the same computational step.

\figureLearning

\section*{Learning Algorithmic Solutions}
\label{sec:exp}
For evaluating the proposed NHC on the three algorithmic requirements R1 -- R3, a diverse set of algorithms was learned and the solutions were tested on their generalization, scaling and abstraction abilities.

The $11$ learned algorithms solve search, plan, addition, sorting, evaluating arithmetic expressions, and sequence retrievals problems.
In Figure~\ref{fig:tasks} all $11$ algorithms are sketched with their pseudocode and examples (more details can be found in the Supplementary Information).

Learning is done in a curriculum learning setup~\supercite{bengio2009curriculum}, where the complexity of presented samples increases with each curriculum level.
During learning, samples up to curriculum level $10$ are considered, with an additional level $11$ that samples from all previous levels.
Generalization and scaling is tested on complexities up to level $1000$.
The direct transfer is tested by transferring the learned solutions to novel problem representations.

\subsection*{Learning Procedure Overview}
The algorithmic modules, encoding the algorithmic solution, are learned via Natural Evolution Strategies (NES)~\supercite{wierstra2014natural}.
In each iteration $t$, a population of $P$ offspring (altered parameters $\theta_t^o$) is generated, and the parameters are updated in the direction of the best performing offspring. 
Parameters are updated based on their fitness, a measurement that scores how well the offspring perform.
Such optimizers do not require differentiable models, giving more freedom to the model design, e.g., using non-differentiable hard memory decisions~\supercite{greve2016evolving} and instantiating the modules freely and flexible.

An update at iteration $t+1$ of the parameters $\theta$ with learning rate $\alpha$ and search distribution variance $\sigma^2$ is performed as $\theta_{t+1} = \theta_t + \alpha\nabla_{\theta_{t}}$ with the sampled NES gradient given as
\begin{align*}
\nabla_{\theta_{t}} \mathbb{E}_{\epsilon \sim N(0, I)}  \left[ f (\theta_t + \sigma\epsilon)  \right] \approx \frac{1}{P\sigma} \sum_{o=1}^{P} f(\theta_t^o)\epsilon_i  \  .
\end{align*}
Hence, parameters are updated based on a performance-weighted sum of the offspring.
Here, the fitness function $f(\cdot)$ scores how many algorithmic steps were done correctly -- if the correct data was manipulated in the correct way at the correct step.
These binary signals for each step are averaged over all steps and all samples in the minibatch to get a scalar fitness value.
This results in a coarse feedback signal and harder learning problem in contrast to gradient based training, where the error backpropagation gives localized feedback to each parameter.

For all algorithms, generalization and scaling (R1) was tested in two ways.
First, testing for scaling to more complex configurations is integrated into our learning procedure, and second, the solutions were tested on complexities far beyond those seen during training. 

A curriculum level is considered solved after a defined number of subsequent iterations with maximum fitness, i.e., with perfect solutions where every bit in every step is correct.
When a new level is unlocked,  samples with higher complexity are presented and hence, if the fitness stays at maximum, the acquired solution scaled to that new complexity.
Learning is only performed in iterations which do not have maximum fitness.

In addition to this built-in generalization evaluation, the learned solutions were tested on complexities far beyond seen during training, i.e., corresponding to curriculum levels $100$, $500$ and $1000$, while being trained only up to level $10$.

\subsection*{Learning Results}
Learning results on all $11$ algorithms are presented in Figure~\ref{fig:learning}, Table~\ref{table:comparison} and the Extended Data Figure~\ref{fig:curves}, where all results are obtained over $15$ runs of each configuration, and Figure~\ref{fig:behavior} illustrates the learned algorithmic behaviour for four algorithms.

In Figure~\ref{fig:learning}, we illustrate in which curriculum levels learning was triggered. 
The top row shows the last level in which training was triggered, the last level were an error occurred, indicating that learning only occurs in the first levels and solutions generalize to subsequent levels, i.e, to higher complexities.
This can be observed for all $11$ algorithms reliably over all runs.
In the bottom row, we investigated in how many iterations learning was observed in each level and in total. 
This highlights the fact that most training happens in the first levels, and subsequent levels only need a few iterations to adapt, if at all.
The total number of learning iterations highlights the efficient training in terms of samples.
This measure provides an indicator for the task complexity, i.e., for sorting $9016$ iterations caused network updates, whereas copying is less challenging and only required $814$ iterations of network updates. 

In Table~\ref{table:comparison} more details of the learning, the generalization and scaling evaluation for R1, and comparison methods are shown.
The \textit{last level} entry shows the last level triggering learning alongside the last level that was solved successfully, highlighting that all runs for all algorithms were able to solve all $11$ training levels while triggering learning only in the first levels.
Next, the table shows the results on testing the solutions on complexities far beyond those seen during training.
Each run was presented $50$ samples from the associated level ($20$ samples for sort levels $500$ and $1000$ due to runtime scaling).

\tableLearning

For the majority of algorithms, all runs generalized perfectly to complexities up to level $1000$.
In the harder tasks, like sorting, some runs fail for perfect generalization, still performing well and the majority of runs also shows perfect generalization.
Note that, a sample from level $1000$ in the sort task requires over $1 \ million$ perfect computational steps to be considered solved.
The performances below $100\%$ for some runs can be explained with the mechanisms of the previous write head.
The model has to learn if the previously read location should be updated and with which information, without explicit feedback on these signals.
Thus, an update mechanism that learned to slightly update the previous location works fine on shorter sequences (like seen during training), but the small changes accumulate on longer sequences and may result in wrong behaviour.
A possible solution would be to add feedback to this signals during training if it can be provided.

Overall, the results summarized in in Figure~\ref{fig:learning} and Table~\ref{table:comparison} show that the solutions learned by the NHC fulfil the algorithmic requirement R1 of generalization.

\figureTransfers
\paragraph{Comparison} 
For comparison we trained four additional models.
First, the DNC~\supercite{graves2016hybrid} model as a state-of-the-art memory-augmented neural.
This model is trained in a supervised setting with backpropagation, i.e., having a much richer and localized learning signal.
It was able to learn some of the baseline algorithms up to level $5$, like addition, copy and reverse, but failed in earlier levels in the remaining tasks, despite being trained for $500k$ iterations.
Notably, the DNC struggled with those tasks requiring to reuse intermediate results or iterating over the data multiple times.

Second, we integrated the DNC into the NHC architecture by replacing the algorithmic modules of the NHC -- controller, memory, bus -- with the original DNC.
This DNC+is+ha model uses the same data modules and is trained like the NHC with NES.
It performs notably better than the DNC, indicating the help of the proposed abstraction mechanisms and the evolutionary training.
Nevertheless, it still is not able to generalize comparable to the NHC and struggles with the same algorithms as the DNC.
More details on these comparisons and their learning are given in the Supplementary Information.

Next, we removed the proposed ancestry linkage (NHC-anc) and the previous location update (NHC-prev) to evaluate their influence.
To counter the removed update head, the NHC-prev model uses two write heads, enabling it to learn a similar update mechanism.
Both models perform better than the DNC+is+ha and are able to learn the majority of algorithms and even achieve perfect generalization in some, strengthening the importance of the evolutionary training and highlighting the influence of the proposed mechanisms.
The performance of the two ablation models depends on the algorithm to learn, i.e., if the algorithm requires the hierarchical knowledge provided by the ancestry linkage or the updating of previously read locations.
Notably, both mechanisms are required to learn the search and plan algorithms. 

These results suggest that the evolutionary training with the proposed abstraction mechanisms and the new memory module are key ingredients for reliably learning algorithmic solutions that generalize and scale, and hence, fulfilling R1.

\figureBehavior
\subsection*{Transfer of the Learned Algorithmic Solutions}
Next, we evaluated the ability to generalize the learned solutions to new problem instantiations, testing the requirements R2 (independence of the data representation) and R3 (independence of the task domain). 
Therefore, the algorithmic solutions were tested on unseen data representations and task domains.
For these transfers, the learned algorithmic modules were used with adapted data modules for the new setups.

In Figure~\ref{fig:transfers} all transfers are illustrated, showing the training setup and the successful transfers.
The transferred solution solved all $11$ curriculum levels in the new setup without triggering learning once, i.e., no single error occurred.

For search and plan, we investigated if the strategy learned in sokoban could be transferred to bigger environments, to a different data representation, to a sliding puzzle problem, and to a robot manipulation task. 
The solutions were learned in $6 \times 6$ environments, and could perfectly solve $8 \times 8$ environments and a changed encoding of the environment, e.g., the penguin represents a wall instead of the agent (see Figure~\ref{fig:transfers}).
In the $3 \times 3$ sliding puzzles, the white space represents empty space on which adjacent tiles can be moved to. 
In the robotic setup, the task is to rearrange the four stacks of boxes from one configuration into another.

The addition, sort and the baseline algorithms -- copy, repeatCopy, reverse, duplicated -- were trained on binary numbers and were successfully transferred to decimal numbers.

The arithmetic algorithm was trained on decimal arithmetic and was transferred to a boolean algebra.
As the atomic operations $[+,-,*,/]$ are part of the data input sequence, the solution is independent from the number of atomic operations, shown by having only two atomic operations \texttt{AND} \& \texttt{OR} in the boolean algebra setup.

\paragraph{Limitations and Assumptions} 
In our transfer experiments, we assumed the same number of operations available for the \textit{ALU} and adapted data modules.
The number of operations needs to be the same as these, together with the control signals from the \textit{Input}, form the abstraction interface between data and algorithmic modules.
This can be relaxed either by including the domain specific operations into the data sequence, as shown with the arithmetic transfer, or by extending the interface between \textit{Bus} and \textit{ALU}.
The learned algorithmic solution is represented by the \textit{Controller}, \textit{Memory}, and \textit{Bus}, which encode the abstract strategies fulfilling R1-R3, building on the data modules implementing the abstract interface.
As the data modules are domain and representation dependent, they need to be relearned or handcrafted for new setups.
Typically learning these modules is less complex than learning a new algorithm as they solve smaller subproblems (and often can be hardcoded), and is a benefit of the modular architecture with its abstraction mechanism and the evolutionary training.

\section*{Conclusion}
\label{sec:conclusion}
A major challenge for intelligent artificial agents is to learn strategies that scale to higher complexities and that can be transferred to new problem instantiations.
We presented a modular architecture for representing and learning such algorithmic solutions that fulfil the three introduced algorithmic requirements: generalization and scaling to arbitrary task configurations and complexities (R1), as well as independence from both, the data representation (R2) and the task domain (R3).
Algorithmic solutions fulfilling R1 -- R3 represent strategies that generalize, scale, and can be transferred to novel problem instantiations, providing a promising building block for intelligent behaviour.

On a diverse set of $11$ algorithms with varying complexities, the proposed NHC was able to reliably learn such algorithmic solutions.
These solutions were successfully tested on complexities far beyond seen during training, involving up to over $1 \ million$ recurrent computational steps without a single bit error, and were transferred to novel data representations and task domains.
Experimental results highlight the importance of the employed abstraction mechanisms, supporting the ablation study results of prior work~\supercite{tanneberg2019learning}, providing a potential building block for intelligent agents to be incorporated in other models.

\paragraph{Discussion}
The modular structure and the information flow of the NHC enable the learning and transfer of algorithmic solutions, and the incorporation of prior knowledge. 
Using NES for learning removes constraints on the modules, allowing for arbitrary instantiations and combinations, and the beneficial use of non-differentiable memories~\supercite{greve2016evolving}.
As the complexity and structure of the algorithmic modules need to be specified, it is an interesting road for future work to learn these in addition, utilizing recent ideas~\supercite{greve2016evolving,merrild2018hyperntm}. 
To speed up computation, parallel models like the neural GPU~\supercite{kaiser2016neural} may be incorporated into the NHC architecture.

The presented work showed how algorithmic solutions with R1 -- R3 can be represented and learned.
Based on this foundation, a challenging and interesting research question is how such algorithms can be learned with less feedback.
The usage of NES allows to provide different kinds of feedback on any connection in the architecture, and on different timescales.
This opens the opportunity to discover new and unexpected strategies, novel algorithms, and may be achieved by incorporating intrinsic motivation~\supercite{oudeyer2009intrinsic,baldassarre2013intrinsically} to explore the space of hidden algorithmic solutions in the model.

\subsection*{Data availability}
Data is generated online during training and the generating methods are provided in the source code. 

\subsection*{Code availability}
The source code of the NHC is available via Code Ocean at \href{https://doi.org/10.24433/CO.6921369.v1}{https://doi.org/10.24433/CO.6921369.v1}. 

%\bibliography{references}
\printbibliography[segment=\therefsegment]
\end{refsegment}
 
\subsection*{Acknowledgements}
This project has received funding from the European Union's Horizon 2020 research and innovation programme under grant agreement No. \#713010 (GOAL-Robots) and No. \#640554 (SKILLS4ROBOTS), and from the Deutsche Forschungsgemeinschaft (DFG, German Research Foundation) under No. \#430054590.
This research was supported by NVIDIA.
We want to thank Kevin O'Regan for inspiring discussions on defining algorithmic solutions.

\subsection*{Author contributions statement}
D.T. conceived the project, designed and implemented the model, conducted the experiments and analysis, created the graphics.
D.T., E.R. and J.P. wrote the manuscript.  

\subsection*{Competing interests}
The authors declare no competing interests.

\clearpage
% max 3000 words
%######################################################################
\begin{refsegment}
\section*{Methods}
\label{sec:methods}
In this section a detailed description of the NHC architecture and its modules is given, the learning procedure is described, the task specific data module instantiations are discussed, and details about the comparison methods are given.

All modules are described with their formal functionality, i.e., the input signals they receive and the output signals they produce, in the form of $Module(\text{inputs}) \longrightarrow \text{output}$.
The information flow is split into data and control signals, denoted with $d$ and $c$ respectively.
In addition to this high level description, details how the output signals are generated are given for each module.

%######################################################################
\subsection*{The Algorithmic Modules}
\label{sec:alg_modules}
The algorithmic modules consists of the \textit{Controller}, \textit{Memory} and \textit{Bus} module and form the core of the NHC architecture.
These modules are learning the algorithmic solution on the control stream and are responsible for the data management in the memory and steer the data manipulation done by the ALU module.
They share similarities with the original DNC~\supercite{graves2016hybrid}, like the temporal linkage and usage vector, but with major changes, e.g., hard decisions for the heads and read modes, two coupled memories, simplified and additional attention mechanisms and more, described in detail next.

%######################################################################
\subsubsection*{Controller}
The controller module receives input from the Input and the signals read from Memory from the previous step.
Additionally feedback signals from the Bus and ALU from the previous step can be activated if desired.
It produces one output signal going to the Memory and Bus modules.
Formally given by
\begin{align*}
Ctr(c_t^{i \rightarrow c} ,\ c_{t-1}^b ,\ c_{t-1}^a ,\ c_{t-1}^m) \longrightarrow c_t^c \ .
\end{align*}
Here we use a single layer of size $L_C$ to learn $c_t^c \in (-1,1)^{L_C}$ at step $t$, given by
\begin{align*}
c_t^c = \text{tanh} (W_c x_c + b_c) \ ,
\end{align*}
with $x_c = [c_t^{i \rightarrow c} ;\ c_{t-1}^b ;\ c_{t-1}^a ;\ c_{t-1}^m]$.
Depending on the task to learn, the feedback signals $c_{t-1}^b$ and $c_{t-1}^a$ can be activated, and more complex instantiations can be used for the controller, like more layers or recurrent networks.

%######################################################################
\subsubsection*{Memory}
The memory module receives signals from the Input and the Controller and is responsible for storing and retrieving information from the two memories.
Therefore it produces two output signals, a data and a control signal, give by
\begin{align*}
Mem(d_t^i ,\ c_t^{i \rightarrow m} ,\ c_t^c) \longrightarrow (d_t^m, c_t^m) \ .
\end{align*}
The memory module has two coupled control and data memories, $M^c$ and $M^d$, which are matrices of size $N \times C$ and $N \times D$ with $N$ locations, $C$ the control memory word size and $D$ the data memory word size.
Multiple write and read heads can be used, where the number of write and read heads is set task dependently to $h_w$ and $h_r$ respectively.

\paragraph{Learnable Interfaces}
As input for all learned layers only the concatenated control signals are used, i.e., $x_m = [c_t^{i \rightarrow m} ;\ c_t^c]$, and the weight matrices $W$ and biases $b$ are the parameters that are learned.

The \textbf{write vectors} $v_t^i \in \Re^{C}$ at step $t$ are the control signals that are stored in $M^c$ via the write heads and given by
\begin{align*}
v_t = W_v x_m + b_v \ ,
\end{align*}
with $v_t$ split into $\{ v_t^i \ | \ \forall i : h_w \}$ for each write head.

The \textbf{previous write vectors} $\hat{v}_t^j \in \Re^C$ at step $t$ are the control signals that are used to update $M^c$ and given by
\begin{align*}
\hat{v}_t = W_{\hat{v}} x_m + b_{\hat{v}} \ ,
\end{align*}
with $\hat{v}_t$ split into $\{ \hat{v}_t^j \ | \ \forall j : h_r \}$ for each read head.

The \textbf{previous erase vectors} $\hat{e}_t^j \in (0,1)^C$ at step $t$ are the control signals used to erase values in $M^c$ and given by
\begin{align*}
\hat{e}_t = \sigma ( W_{\hat{e}} x_m + b_{\hat{e}} ) \ ,
\end{align*}
where $\sigma(\cdot)$ is the logistic sigmoid function and $\hat{e}_t$ is split into $\{ \hat{e}_t^j \ | \ \forall j : h_r \}$ for each read head.

The \textbf{previous write gate} $\hat{g}_t \in \{0,1\}^{h_r}$ at step $t$ determines if the memory $M^c$ is updated with $\hat{v}_t^j$ and $\hat{e}_t^j$, given by
\begin{align*}
\hat{g}_t = \text{H} ( W_{\hat{g}} x_m + b_{\hat{g}} )  \ ,
\end{align*}
where $H(\cdot)$ is the heavyside step function.

The \textbf{read modes} $m_t^j \in \{0,1\}^{h_r + 4 h_w}$ at step $t$ are the control signals that determine which attention mechanism is used to read from the memory, and is given by
\begin{align*}
m_t = W_m x_m + b_m \ ,
\end{align*}
with $m_t$ split into $\{ \text{onehot}(m_t^j) \ | \ \forall j : h_r \}$ and $\text{onehot}(x) = \{x_k = 1 \ \text{if} \ x_k = \text{max}(x) \ ,\ x_k = 0 \ \text{else}\}$.

The \textbf{free gates} $f_t^w \in \{ 0,1 \}^{h_w}$ and $f_t^r \in \{ 0,1 \}^{h_r}$ at step $t$ determine if locations written to and read from can be freed after interaction, and are given by
\begin{align*}
f_t^w = \text{H} ( W_{f^w} x_m + b_{f^w} ) \quad \text{and} \quad f_t^r = \text{H} ( W_{f^r} x_m + b_{f^r} ) \ .
\end{align*}

These are all learned parameters of the memory module that define the interfaces to manipulate the memory.

\paragraph{Writing and Reading}
Given the learned interface described before and the write $w_t^i$ and read $r_t^j$ head locations, information is stored and retrieved from memory as follows. 

Writing $v_t^i$ to location $w_t^i$ in $M^c$ at step $t$ is done via
\begin{align*}
M_t^c = M_{t-1}^c \circ (E - w_t^i 1^{\top}) + w_t^i {v_t^i}^{\top} \ ,
\end{align*} 
where $\circ (\cdot)$ denotes element-wise multiplication and $E$ is a matrix of ones of the same size as $M^c$.

Writing $d_t^i$ to location $w_t^i$ in $M^d$ at step $t$ is done via
\begin{align*}
M_t^d = M_{t-1}^d \circ (E - w_t^i 1^{\top}) + w_t^i {d_t^i}^{\top} \ ,
\end{align*} 
here $E$ is a matrix of ones of the same size as $M^d$.
Note that the same write location $w_t^i$ is used to couple the control and data memories. 

Updating the previously read location $r_{t-1}^j$ in $M^c$ is done via
\begin{align*}
M_t^c = M_{t-1}^c \circ (E - \hat{g}_t^j r_{t-1}^j {{\hat{e}}_t}^j{^{\top}} ) + \hat{g}_t^j r_{t-1}^j {{\hat{v}}_t}^j{^{\top}} \ ,
\end{align*} 
where $E$ is a matrix of ones of the same size as $M^c$.
If the previous write gate $\hat{g}_t^j = 0$ no update is performed, and with $\hat{g}_t^j = 1$ the previously read location $r_{t-1}^j$ is erased with ${{\hat{e}}_t}^j$ and ${{\hat{v}}_t}^j$ is written to it.

Reading from memory is done via the read locations $r_t^j$ used on both memories to obtain the data and control output of the memory module via
\begin{align*}
d_t^{m,j} = {M_t^d}^{\top} r_t^j \quad \text{and} \quad c_t^{m,j} = {M_t^c}^{\top} r_t^j \ ,
\end{align*} 
and are concatenated for the final memory module output $d_t^m = [ d_t^{m,1}; \dots ; d_t^{m,h_r}]$ and $c_t^m = [ c_t^{m,1}; \dots ; c_t^{m,h_r}]$.

Next, how to obtain the head locations is described in detail.
 
\paragraph{Head Locations}
The write and read heads locations, $w_t^i \in \{0,1\}^N$ and $r_t^j \in \{0,1\}^N$, are hard decisions, i.e., onehot encoded vectors, where exactly one location is written to or read from respectively.
To determine the write head locations $w_t^i$, i.e., the memory locations for writing to, a simplified dynamic memory allocation scheme from the DNC is used.
It is based on a \textit{free list} memory allocation scheme, where a linked list is used to maintain the available memory locations.
Here, a usage vector $u_t \in \{0,1\}^N$ indicates  which memory locations are currently used, with $u_0 = 0$ and updated in each step with the written $w_t^i$ and read $r_t^j$ locations via
\begin{align*}
u_t = u_{t-1} + (1-f_t^{w,i}) w_t^i \quad \text{and} \quad u_t = u_{t-1} (1 - f_t^{r,j} r_t^j) \ ,
\end{align*} 
with the free gates $f_t^{w,i}$ and $f_t^{r,j}$ determining if the write location is marked as used and if the read location can be freed respectively.
Due to this dynamic allocation scheme, the model is independent from the size of the memory, i.e., can be trained and later used with different sized memories.
To obtain the write location $w_t^i$, the memory locations are ordered by their usage $u_t$, and $w_t^i$ is set to the first entry in this list -- the first unused location is used to write to.

Read head locations $r_t^j$ are determined by the active read mode given by $m_t^j \in \{0,1\}^{h_r + 4 h_w}$, i.e, only one mode can be active.
There are three main attentions implemented for reading from memory, \textit{HALT}, \textit{temporal linkage} and \textit{ancestry linkage}.
The total number of available read modes is $h_r + 4 h_w$ as \textit{HALT} is depended on the number of read heads and both linkages can be used in two directions for each write head.

The \textit{HALT} attentions are used to read the previously read locations again.
When multiple read heads are used, each head can read its own last location or the locations from the other read heads, e.g., with three read heads, each head has three \textit{HALT} attentions (\textit{H1,H2,H3}).

The \textit{temporal linkage} attention is used to read locations in the order they were written, either in forward or backward direction.
This mechanism enables the architecture to retrieve sequences, or parts of sequences, in the order they were presented, or in reversed order.
Here, we use a simplified version of the mechanism from the DNC.
As our architecture uses hard decisions for the heads locations, the linkages can be stored more efficiently in $N$-dimensional vectors, in contrast to a $N \times N$ matrices in the DNC.
Each temporal linkage vector $L^{T,i}$ stores the order of write locations for one write head, updated at step $t$ via
\begin{align*}
L_t^{T,i} = L_{t-1}^{T,i} \circ (1 - w_{t-1}^i) + \tilde{w}_t^i w_{t-1}^i \ ,
\end{align*} 
where $\tilde{w}_t^i = \text{argmax}(w_t^i)$.
The temporal linkage mechanism can be used in two directions.
Either move the read head in the order of which the locations were written, or in reversed order -- resulting in two read modes, backward (B) and forward (F), per write head for each read head, given by
\begin{align*}
B:& \ r_t^j = \text{I}(L_t^{T,i} \ , \ \tilde{r}_{t-1}^j) \quad \text{and}
\\
F:& \ r_t^j = \text{onehot}( L_t^{T,i} \circ r_{t-1}^j ) \ ,
\end{align*} 
where $r_{t-1}^i$ is the previously read location, $\tilde{r}_{t-1}^j = \text{argmax}(r_{t-1}^i)$ and $\text{I}(x,y) = \{x_k = 1 \ \text{if} \ x_k = y \ ,\ x_k = 0 \ \text{else}\}$.
When a location is freed through the free gates, the location is removed from the linkage such that it remains a linked list.

The \textit{ancestry linkage} also uses $N$-dimensional vectors to store relations between memory locations.
While the temporal linkage stores information about the order of which locations were written to, the ancestry linkage stores information about which memory locations were read before a location was written -- captures a form of \textit{usage} or hierarchical relation instead of temporal relation.
Each ancestry linkage vector $L^{A,i,j}$ stores which location $r_{t-1}^j$ was read before location $w_t^i$ was written, and is updated at step $t$ via
\begin{align*}
L_t^{A,i,j} = L_{t-1}^{A,i,j} \circ (1 - w_t^i) + \tilde{r}_{t-1}^j w_t^i \ ,
\end{align*} 
where $r_{t-1}^j$ is the previously read location and $\tilde{r}_{t-1}^j = \text{argmax}(r_{t-1}^j)$.
The ancestry linkage mechanism can also be used in two directions, to either move the read head to \textit{parent} (P) location or the \textit{child} (C) location.
This results in two modes per write head for each read head, given by
\begin{align*}
P:& \ r_t^j = \text{onehot}( L_t^{A,i,j} \circ r_{t-1}^j ) \quad \text{and}
\\
C:& \ r_t^j = \text{onehot}( \text{I}( L_t^{A,i,j} \ , \ \tilde{r}_{t-1}^j ) \circ \text{h}_t ) \ ,
\end{align*} 
where $\text{h}_t$ is a $N$-dimensional vector storing for each location the step $t$ when it was written.
A location can be read multiple times, thus it can have multiple \textit{children}.
But as we need a single location to read, the C mode returns the location that was written to the latest when $r_{t-1}^j$ was read, i.e., the newest child.
This is implemented with the history vector $\text{h}_t$.
When a location is freed through the free gates, the location is removed from the linkage and its children are attached to its parent.

%######################################################################
\subsubsection*{Bus}
The Bus module is responsible to generate the control signal that indicates how the ALU module should manipulate the data stream, i.e., which action or operation to perform.
Therefore, it receives the control signal from the Controller and the Input as well as the output from the memory signal, given by
\begin{align*}
Bus(c_t^{i \rightarrow b} ,\ c_t^c ,\ c_t^m) \longrightarrow c_t^b \ .
\end{align*}

Here we use a single layer of size $L_B$ to learn $c_t^b \in \{ 0,1 \}^{L_B}$ at step $t$, given by
\begin{align*}
c_t^b = \text{onehot} (W_b x_b + b_b) \ ,
\end{align*}
with $x_b = [c_t^{i \rightarrow b} ;\ c_t^c ;\ c_t^m]$.

%######################################################################
\subsection*{Learning Procedure}
Learning the algorithmic modules, and hence the algorithmic solution, is done using Natural Evolution Strategies (NES)~\supercite{wierstra2014natural}.
NES is a blackbox optimizer that does not require differentiable models, giving more freedom to the model design, e.g., the hard attention mechanisms are not differentiable and the data modules can be instantiated arbitrarily.
Recent research showed that NES and related approaches like Random Search~\supercite{mania2018} or NEAT~\supercite{stanley2002evolving} are powerful alternatives to gradient based optimization in reinforcement learning.
They are easier to implement and scale, perform better with sparse rewards and credit assignment over long time scales, have fewer hyperparameters~\supercite{salimans2017evolution} and were used to train memory-augmented networks~\supercite{greve2016evolving,merrild2018hyperntm,khadka2019neuroevolution}.

NES updates a search distribution of the parameters to be learned by following the natural gradient towards regions of higher fitness using a population $P$ of offspring $o$ (altered parameters) for exploration.
The performance of an offspring $o$ is measured with one scalar value summarized over all samples $N$ in the mini-batch and over all computational steps $T_{\text{max}}$ of each sample, with sparse binary signals for each step -- framing a challenging learning problem albeit given the sequence.
Let $\theta$ be the parameters to be learned -- the weight matrices and biases in the three algorithmic modules $\theta = [W_c ; b_c ; W_v ; b_v ; W_{\tilde{v}} ; b_{\tilde{v}} ; W_{\tilde{e}} ; b_{\tilde{e}} ; W_{\tilde{g}} ; b_{\tilde{g}} ; W_m ; b_m ; W_{f^w} ; b_{f^w} ; W_{f^r} ; \linebreak b_{f^r} ; W_b ; b_b]$ -- and using an isotropic multivariate Gaussian search distribution with fixed variance $\sigma^2$, the stochastic natural gradient at iteration $t$ is given by
\begin{align*}
\nabla_{\theta_{t}} \mathbb{E}_{\epsilon \sim N(0, I)}  \left[ u (\theta_t + \sigma\epsilon)  \right] \approx \frac{1}{P\sigma} \sum_{o=1}^{P} u(\theta_t^o)\epsilon_i  \  ,
\end{align*}
where $P$ is the population size and $u(\cdot)$ is the rank transformed fitness $f(\cdot)$~\supercite{wierstra2014natural}.
With a learning rate $\alpha$, the parameters are updated at iteration $t$ by
\begin{align*}
\theta_{t+1} = \theta_t + \frac{\alpha}{P\sigma} \sum_{o=1}^{P} u(\theta_t^o)\epsilon_i  \  .
\end{align*}
For all experiments the fitness function is defined for $N$ samples as $f(\theta_t^o) = 1 / N \sum_{s}^{N} f_s(\theta_t^o)$ with
\begin{align*}
f_s(\theta_t^o) = \frac{1}{T_{\text{max}}} \sum_{k=1}^{T_e} \delta(d_k^m - \bar{d}_k^m) + \delta(c_k^b - \bar{c}_k^b) m(c_k^b) \ 
\end{align*}
to evaluate the offspring parameters $\theta_t^o$ on one sample $s$.
Here, $\delta(x) = \{1 \ \text{if} \ x = 0, 0 \ \text{else} \}$ gives sparse binary reward if the two signals are equal or not, where $d_k^m$ is the data output from the memory, $c_k^b$ the control output from the Bus, and $\bar{d}_k^m$ and $\bar{c}_k^b$ the true values respectively.
Thus, reward is given for choosing the correct data and operation for the ALU in each step.
Note, there is no feedback on memory access, only on the output, i.e., where, when and how to write and read has to be learned without explicit feedback.
The stepwise signals are summed up until the first mistake occurs ($T_e$) or until the maximum length of the sample $T_{\text{max}}$, and is normalized with $1/T_{\text{max}}$, i.e., $f(\theta_t^o)$ measures the fraction of subsequently correct algorithmic steps.
To encourage strong operation choices, the operation reward is multiplied with the margin penalty 
\begin{align*}
m(c_k^b) = clip \big( \frac{\tilde{c}^1 / \tilde{c}^2 - 1}{m_{\text{max}}} \ , \ 0 \ , \ 1 \big) \ ,
\end{align*}
where $\tilde{c}^1 , \tilde{c}^2$ are first and second largest values of $c_k^b$, i.e., the chosen operation and the runner up, and $m_{\text{max}}$ is a chosen percentage indicating how much bigger the chosen action should be.
Note that this penalty is only considered if the operation is already correct.

For robustness and learning efficiency, weight decay for regularization~\supercite{krogh1992simple} and automatic restarts of runs stuck in local optima are used~\supercite{wierstra2014natural}.
This restarting can be seen as another level of evolution, where some lineages die out.
Another way of dealing with early converged or stuck lineages is to add intrinsic motivation signals like novelty, that help to get attracted by another local optima, as in NSRA-ES~\supercite{conti2018improving}.
In the experiments however, we found that within our setting, restarting -- or having an additional \textit{survival of the fittest} on the lineages -- was more effective in terms of training time.

The algorithmic solutions are learned in a curriculum learning setup~\supercite{bengio2009curriculum} with sampling from old lessons to prevent unlearning and to foster generalization.
Furthermore, we created \textit{bad memories}, a learning from mistakes strategy, similar to the idea of AdaBoost~\supercite{freund1997decision}, which samples previously failed samples to encourage focusing on the hard cases.
This can also be seen as a form of experience replay~\supercite{mnih2015human,lin1992self}, but only using the initial input data to the model, not the full generated sequences.
Bad memories were initially developed for training the data-dependent modules to ensure their robustness and $100\%$ accuracy, which is crucial to learn algorithmic solutions.
If the individual modules do not have $100\%$ accuracy, no stable algorithmic solution can be learned even if the algorithmic modules are doing the correct computations. 
For example, if one module has an accuracy of $99\%$, the $1\%$ error prevents learning an algorithmic solution that works \textit{always}.
This problem is even reinforced as the proposed model is an output-input architecture that works over multiple computation steps using its own output as the new input -- meaning the overall accuracy drops to $36.6\%$ for $100$ computation steps.
Therefore using the bad memories strategy, and thus focusing on the mistakes, helps significantly in achieving robust results when learning the modules, enabling the learning of algorithmic solutions.

\paragraph{Experimental Setup} In all experiments, the hyperparameters were set to: batch size $N = 32$, population size $P = 20$, learning rate $\alpha = 0.01$, search distribution exploration $\sigma = 0.1$, weight decay $\lambda = 0.9995$, action margin $m_{\text{max}} = 0.1$, max iterations = $20.000$, restart iterations = $2.000$.
In each batch, $33\%$ of the samples were drawn from previous levels and another $33\%$ were drawn from the bad memories buffer, which stores the last $200$ mistakes.
A curriculum level is considered solved when $750$ subsequent iteration are perfectly solved, i.e., no single mistake in any sample, any step, any bit, that is $24.000$ perfectly solved samples.
In levels were training was triggered, the required subsequent perfect iterations are doubled, i.e., $48.000$ perfectly solved samples.
Whenever an iteration achieves maximum fitness, no learning is triggered, i.e., no parameter update is performed.

%######################################################################
\subsection*{The Modules Instantiations}
\label{sec:task_modules}
The preceding sections described the design and functionality of the algorithmic modules in general.
Here, the used instantiations and parameters for the experiments are presented as well as the data modules and their task-dependent instantiations.

\paragraph{Algorithmic Modules}
In all experiments, the Controller size was set to $L_C = 6$ and the control memory word size $C = 4$.
All tasks use one write head, $h_w = 1$, and the number of read heads is $h_r = 1$ for the four search \& plan, and the four copy tasks, $h_r = 2$ for addition and sort  and $h_r = 3$ for the arithmetic task.
The data memory word size $D$ and the Bus size $L_B$ are set by the task, as each task has a different data representation ($D$) and a different amount of available operations for the ALU ($L_B$).
In all tasks the ALU to Controller feedback ($c_{t-1}^a$) was activated, except for the four copy tasks as the ALU has no functionality there.
The free gates were activated for the four copy tasks and the arithmetic task.
In total, depending on the algorithm to learn, this results in $300$-$650$ trainable parameters in the algorithmic modules.

\paragraph{Input}
The first data dependent module is the Input module.
That is the interface to receive data and provide control signals.
It receives the data input from the outside $d_t^{in}$ as well as the data output from the ALU from the previous computational step $d_{t-1}^{out}$, formally given as
\begin{align*}
In(d_t^{in} ,\ d_{t-1}^{out}) \longrightarrow (d_t^i ,\ c_t^{i \rightarrow c} ,\ c_t^{i \rightarrow m} ,\ c_t^{i \rightarrow b}) \ .
\end{align*}
The main functionality is to generate task related control signals, data preprocessing if applicable and determining to stop.
The control signals $c_t^{i \rightarrow c} , c_t^{i \rightarrow m} , c_t^{i \rightarrow b}$ can be different to provide different signals to the Controller, Memory and Bus, but can also share the same information.
The data $d_t^i$ is forwarded to the memory module with or without preprocessing, depending on the task.

\paragraph{ALU}
The arithmetic logic unit (ALU) module is responsible for data manipulation.
It receives the read data from the memory and the operation to apply on these from the Bus to produce the next data output alongside control signals via
\begin{align*}
ALU(d_t^m ,\ c_{t}^{b}) \longrightarrow (d_t^{out}, c_t^{a}) \ .
\end{align*}
This module implements elemental operations for each task such that the algorithmic solution can be learned by applying the correct operation on the correct data in the correct step.

Both data modules can be instantiated arbitrarily due to the NES approach for learning the algorithmic solution.
They can also be trained from data beforehand or be hardcoded if possible.
In the experiments, we tested both variations and details for each algorithm are given in the Supplementary Information.

\printbibliography[segment=\therefsegment, keyword=method, heading=subbibliography]
\end{refsegment}

\begin{refsegment}
\clearpage
\pagenumbering{roman}
\section*{Supplementary Information}
Supplementary Information to the paper \textit{Evolutionary Training and Abstraction Yields Algorithmic Generalization of Neural Computers}.

\section*{Algorithms to Learn}
In this section additional information for the different algorithms that are learned and the modules instantiations are given.

\subsection*{Search \& Plan}
In the Search \& Plan tasks, the goal is to reach a given goal from a starting state (search), and to generate a path between them utilizing the search result (plan).
Therefore, a breadth-first-search algorithm with additionally backtracking is learned.
Given a start and goal state ($s_0$ and $g$), the model has to learn to implicitly build a search tree by applying all available actions on a state and then move on to the next unexplored state until the goal state is reached.
For the planning tasks, after reaching the goal state, the model has to output the sequence of states from goal to start, encoding the sequence of actions to solve the given planning problem.
For the extended versions (search+ and plan+), the available actions for each state are state dependent such that only actions that have an effect in this state should be applied.
This increases the complexity of the learned algorithm, but the resulting search trees after learning are smaller and, hence, search+ and plan+ more efficient.

As training domain, the gridworld game sokoban is used, where an agent can move in four directions -- move up ($U$), right ($R$), down ($D$), left ($L$) -- and an additional \texttt{nop} operation($N$) that leaves the data unchanged, resulting in $5$ operations.
The world consists of empty spaces that can be entered, walls that block movement and boxes that can be pushed onto adjacent spaces.
Figure~\ref{fig:behavior} shows a learned solution, where the agent is visualized as a penguin, empty space is water, boxes are icebergs and walls are ice floes.

The curriculum level complexity is the number of fully explored nodes, i.e., for level $3$, three nodes have to be explored.
See Figure~\ref{fig:tasks} for an example for the extended version from level $3$ alongside the pseudocode of the algorithm to learn. 

For learning, we use a sokoban world of size $6\times6$ that is enclosed by walls.
A world is represented with binary vectors and four-dimensional one-hot encodings for each position, resulting in $144$-dimensional data words, and thus $D = 144$.
The configuration of each world -- inner walls, boxes and agent position -- is sampled randomly.
Each world is generated by sampling uniformly the number of inner walls from $[0,2]$ and boxes from $[1,5]$.
The positions of these walls, boxes and the position of the agent are sampled uniformly from the empty spaces.

\paragraph{Input}
The Input module produces control signals indicating in which phase the algorithm is -- building the search tree, goal reached and backtracking the solution.
The signal is created as $c_t^{p} = [c_{t}^{[1]} ; c_{t}^{[2]} ; c_{t}^{[3]} ; c_{t}^{[4]}]$, with
\begin{align*}
c_{t}^{[1]} &= \text{max}(0, (1 - E_t) - c_{t-1}^{[2]} ) \  , \\
c_{t}^{[2]} &= \text{min}(1, E_t + c_{t-1}^{[2]} ) \ , \\ 
c_{t}^{[3]} &= (1 - E_t) p_t^{[2]} \ , \\ 
c_{t}^{[4]} &= E_t p_{t-1}^{[2]} \ .
\end{align*}
The signal $E_t$ is a learned equality function using differential rectifier units as inductive bias~\supercite{weyde2018feed} and consists of a feedforward network with $10$ hidden units and \texttt{leaky-ReLU} activation, trained in a supervised setting with cross-entropy loss.
This signal indicates if two given states are equal and is used to identify the goal and initial state.
The four signals $c_{t}^{[x]}$ are created using this information and represent the state of the algorithms, where $c^{[1]} = 1$ during building the search tree, $c^{[2]} = 1$ when the goal was found, $c^{[3]} = 1$ during the backtracking, and $c^{[4]} = 1$ when backtracking reached the initial state.
The algorithm stops if $c_{t}^{[4]} = 1$ for the planning tasks, and if $c_{t}^{[2]} = 1$ for the search tasks, as search only uses the two first signals.

For the extended search and plan tasks, the Input module additionally learned an action mask $a$.
This binary action mask indicates which operations are applicable in a given state and which not, and is learned with a feedforward network with $256$ hidden units and \texttt{leaky-ReLU} activation.

Using these learned signals, the outputs of the Input module are given by
\begin{align*}
d_t^i &= d_t^{in} \quad \text{if} \quad t = 0 \ , \quad d_{t-1}^{out} \quad \text{else} \ , \\
c_t^{i \rightarrow c} &= c_t^{p} \ , \\
c_t^{i \rightarrow b} &= \ c_t^{i \rightarrow c}, \\
c_t^{i \rightarrow m} &= c_t^{i \rightarrow c} \quad \text{or} \quad c_t^{i \rightarrow m} = [a ; c_t^{p}] \ \text{(extended)} \ . 
\end{align*}

\paragraph{ALU}
The ALU learns to apply the available operations on the data, i.e., it encodes an action model $A$ by learning preconditions and effects, and outputs the new data together.
There are four available operations in sokoban and the sliding block puzzle domain, i.e., move up ($U$), right ($R$), down ($D$), left ($L$), and a fifth \texttt{nop} operation is added that does not change the data ($N$).
In the robotic manipulation domain, the available actions are the four locations on which objects can be stacked, e.g., the action pos1 encodes to move the gripper to the position and place the grasped object on top, or to pick up the top object if no object is grasped. The maximum stacking height is 3 boxes, resulting in a discrete representation of the object configuration with a $3 \times 4$ grid.
Thus, the ALU for the search and plan algorithms has five operations and, hence, $L_B = 5$.

For this task, the ALU internally uses three submodules -- one does dimensional reduction on the data, one applies the operation to manipulate the lower-dimensional data, and one combines the manipulated and original data for the final output. 
Each submodule consists of feedforward networks with $1: 500$, $2: ([128,64], [64,64])$, $3: ([500,500], [500,250])$ hidden units to produce the data and control stream respectively.
All networks use \texttt{leaky-ReLU} activation and are trained with cross-entropy loss in a supervised setting.
The output of the ALU is therefore defined as
\begin{align*}
d_t^{out} &= A(d_t^m ,\ c_{t}^{b}) \ , \\
c_t^{a} &= [l, 1-l, n] \ ,
\end{align*}
where $d_t^m$ is the read state, $c_{t}^{b}$ the operation to apply, $l = 1$ if the applied operation was the last for this state, and $n = 1$ if the \texttt{nop} operation was used.

A more detailed description of these submodules and the search \& plan specific data modules can be found in the predecessor model~\supercite{tanneberg2019learning} used for symbolic planning tasks.

\subsection*{Addition}
In the addition task, two numbers have to be added.
The two numbers $a,b$ are presented subsequently in big-endian order.
To solve the task, the model has to learn to read the two numbers correctly aligned, add the corresponding bits and remember the carry for the next step.
Therefore, $3$ operations are available, adding two given bits without ($A$) and with carry bit ($C$), and a \texttt{nop} operation ($N$).

Curriculum level complexities are defined as the bit length of the numbers $a,b$, i.e., in level $3$ numbers of length $3$ have to be added, and $a,b$ have an additional leading $0$.
For training, $a,b$ were randomly generated binary numbers and thus $D = 1$.

\paragraph{Input}
The Input module produces control signals indicating if $a$ or $b$ is presented, or if addition should be done.
Thus, the module output is given by
\begin{align*}
d_t^i &= d_t^{in} \quad \text{if} \quad t \leq 2T_l \ , \quad 0 \quad \text{else} \ , \\
c_t^{i \rightarrow c} = c_t^{i \rightarrow m} = c_t^{i \rightarrow b} &= [c_a ; c_b ; c_s]  \ ,
\end{align*}
where $T_l$ is the the length of $a$ and $b$, $c_a = 1$ and $c_b = 1$ when $a$ or $b$ are presented respectively, and $c_s = 1$ during the addition phase.
The algorithm stops after $T_l$ steps with $c_s = 1$.

\paragraph{ALU}
There are $3$ operations for the ALU module here, adding two given bits without ($A$) and with carry bit ($C$), and a \texttt{nop} operation ($N$), and hence, $L_B = 3$.
The data input consists of two bits $d_t^m = [v_1 ; v_2]$, read from the memory with the two read heads.
The outputs of the ALU are given by
\begin{align*}
d_t^{out} &= v_1 + v_2 + carry \ , \\
c_t^{a} &= [c, 1-c, n] \ ,
\end{align*}
where $carry = 1$ if the operation $C$ is chosen, $c = 1$ if the current operation produced a carry bit, and $n = 1$ if the \texttt{nop} operation was used.

\subsection*{Sort}
In the sort task the model is given an unordered list of objects and has to output the objects in order.
There are $3$ operations, comparing two objects ($C$), skipping the current objects ($S$), and outputting one object ($O$).
To solve the task, the model has to iterate over the sequence of objects and output the \textit{smallest} object in each iteration.
Note, it does not need to be the smallest, if the sequence should be ordered in descending order for example, it outputs the largest.
The order is defined by the compare operation implemented in the ALU module, the learned algorithm can therefore order any sequence in any order.

Curriculum level complexities are defined as the length of the sequence to order, e.g., in level $3$ sequences of length $4$ have to be sorted.
For training, the sequences consist of randomly generated $8$ bit binary numbers ($D = 8$) and the ALU uses $lessEqual$ as $compare$ function.

\paragraph{Input}
The Input module generates control signals indicating if the unordered sequence is presented or if sorting should be done.
Thus, the module output is given by
\begin{align*}
d_t^i &= d_t^{in} \quad \text{if} \quad t \leq T_s \ , \quad 0 \quad \text{else} \ , \\
c_t^{i \rightarrow c} = c_t^{i \rightarrow m} = c_t^{i \rightarrow b} &= [c_f ; c_e ; c_l ; c_s]  \ ,
\end{align*}
where $T_s$ is the length of the sequence, $c_f = 1$ if the first object in the sequence is presented, $c_l = 1$ for the last object, $c_e = 1$ for all other objects, and $c_s = 1$ during the sort phase.
The algorithm stops if the output operation was used $T_s$ times during the sort phase.

\paragraph{ALU}
The ALU has $3$ operations, a compare operation ($C$) that compares two objects, a skip operation ($S$), and a output operation ($O$) to mark one object for output, so $L_B = 3$.
The data input consists of two objects $d_t^m = [o_1 ; o_2]$, read from the memory with the two read heads.
The outputs of the ALU are given by
\begin{align*}
d_t^{out} &= o_1 \ , \\
c_t^{a} &= [c, 1-c, s, o] \ ,
\end{align*}
where $c = 1$ if $compare(o_1, o_2) = true$, $s = 1$ if the skip operation was used, and $o = 1$ if the output operation was used.
For training, sequences consisted of binary numbers and $compare(o_1, o_2) = o_1 \leq o_2$, i.e., sorting numbers in ascending order.

\subsection*{Arithmetic}
In the arithmetic task the model is given a sequence that encodes an arithmetic expression in postfix notation, e.g., $3 + 5 * 6$ is presented as $5 \ 6 * 3 +$.
There are $2$ operations, a calculation operation ($C$) that calculates a given atomic operation, and a read operation ($R$).
To solve the task, the model has to essentially learn to emulate a stack.
It iterates over the input sequence and learns that numbers need to be stored on the stack, and if an atomic operation is presented, takes the two most recent numbers from the stack to combines them accordingly until the input sequence has finished.
As the atomic operations $[+,-,*,/]$ are part of the input sequence, the solution is independent from the amount of different atomic operations, and the ALU has the two described operations.

For training, the sequences consist of arithmetic expression with the four atomic operations $[+,-,*,/]$, where a modulo $10.000$ operation was applied to atomic results for numeric stability, and numbers in the input sequence were drawn from $[1,10]$, and hence $D = 1$.
Training sequences are generated randomly, with uniformly sampled atomic operations and numbers.
Curriculum level complexities are defined as the number of atomic operations, i.e., $5 \ 6 * 3 +$ is an example for level $2$.

\paragraph{Input}
The Input module provides controls signals indicating if the current data word is a value or an atomic operation.
Therefore, the module outputs are given by
\begin{align*}
d_t^i &= d_t^{in} \quad \text{if} \quad c_{t-1}^{i \rightarrow c} = [1,0] \ , \quad d_{t-1}^{out} \quad \text{else} \ , \\
c_t^{i \rightarrow c} &= c_t^{i \rightarrow m} = c_t^{i \rightarrow b} = [c_v ; c_a]  \ ,
\end{align*}
where $c_v = 1$ when $d_t^i$ is a value (e.g., a number), and $c_a = 1$ is $d_t^i$ is an atomic operation.
The algorithm stops if the last atomic operation was presented.

\paragraph{ALU}
The ALU module has $2$ operations, a calculation operation ($C$) that calculates a given atomic operation, and a read operation ($R$), and hence $L_B = 2$.
The data input consists of three values $d_t^m = [d_1 ; d_2 ; d_3]$, read from the memory with the three read heads.
The module outputs are then given by
\begin{align*}
d_t^{out} &= d_1(d_2,d_3)  \quad \text{if} \quad C \ , \quad d_2 \quad \text{else} \ , \\
c_t^{a} &= c_t^b \ ,
\end{align*}
where $c_t^b$ is the signal coming from the Bus module, indicating which operation to apply, i.e., here $c_t^b = [C ; R]$ with $C = 1$ is an atomic operation should be used, or $R = 1$ if not.
If $C = 1$, the first data word $d_1$ read from the memory is interpreted as the atomic operation and is applied on $d_2$ and $d_3$, e.g., if $d_t^m = [+ ; 4 ; 5]$ then $d_t^{out} = d_1(d_2,d_3) = 4 + 5 = 9$.
The four atomic operations $[+,-,*,/]$ are encoded as $[-1, -2, -3, -4]$ in the input, as the input sequence can only contain positive numbers in the used setup -- the learned algorithm is independent from that choice.

\subsection*{Copy, RepeatCopy, Reverse, Duplicated}
In the four baseline tasks -- Copy, RepeatCopy, Reverse, Duplicated -- there is no data manipulation.
For solving these tasks, the model has to learn the proper data management.
There are $2$ ALU operations to \textit{mark} the data ($O$ and $M$), which do not alter the data.

In the \textit{copy} task, the model is presented a sequence of objects $L$ and has to output the same sequence of objects, i.e.,  $L = [x_1,\dots,x_n] \longrightarrow [x_1,\dots,x_n]$.
Therefore, it needs to learn to iterate over the data in the presented order.

In the \textit{repeatCopy} task, the model is also presented a sequence of objects $L$ and has to output the sequence $c$ times, i.e.,  $L = [x_1,\dots,x_n],c \longrightarrow [x_1,\dots,x_n,x_1,\dots,x_n,\dots]$.
Here, it needs to learn to iterate multiple times over the data, requiring to jump back to the start of the sequence.

In the \textit{reverse} task, the model is presented a sequence of objects $L$ and has to output the sequence in reversed order, i.e.,  $L = [x_1,\dots,x_n] \longrightarrow [x_n,\dots,x_1]$. 
Solving requires to learn to iterate over the data in reversed order of presentation.

In a remove duplicates (\textit{duplicated}) task, the model is presented a sequence of objects $L$ with duplicates of each object and has to output the sequence without these duplicates, i.e.,  $L = [x_1,x_1,x_1,x_2,x_2,x_2\dots,x_n,x_n,x_n] \longrightarrow [x_1,\dots,x_n]$.
In order to solve this task, the model has to learn during the presentation of the input sequence which data \textit{to ignore}, while getting only feedback for outputting the sequence without duplicates.

Curriculum level complexities are defined as the number of objects in the sequence, and as the number of copies or duplicates for the repeatCopy and duplicated tasks respectively.
For training, objects were random binary vectors of length $6$ ($D = 6$).

\paragraph{Input}
The Input module provides control signals indicating if objects are presented or output should be done.
For the copy, repeatCopy and reverse tasks, the modules outputs are given by
\begin{align*}
d_t^i &= d_t^{in} \quad \text{if} \quad t \leq T_L \ , \quad 0 \quad \text{else} \ , \\
c_t^{i \rightarrow c} &= c_t^{i \rightarrow m} = c_t^{i \rightarrow b} = [c_f ; c_i ; c_l ; c_o]  \ ,
\end{align*}
where $T_L$ is the length of the sequence $L$, $c_f = 1$ if the first object is presented, $c_l = 1$ for the last object, $c_i = 1$ for the remaining objects, and $c_o = 1$ indicating the output phase.
The algorithm stops, if the $M$ action was used $cT_l$ times in the output phase, with $c = 1$ for copy and reverse.

For the duplicated task, the outputs are given by
\begin{align*}
d_t^i &= d_t^{in} \quad \text{if} \quad t \leq T_L \ , \quad 0 \quad \text{else} \ , \\
c_t^{i \rightarrow c} &= c_t^{i \rightarrow m} = c_t^{i \rightarrow b} = [c_f ; c_i ; c_o]  \ ,
\end{align*},
where $T_L$ is the length of the sequence $L$ including the duplicates, $c_f = 1$ when an object is presented the first time, $c_i = 1$ for the remaining times, and $c_o = 1$ indicating the output phase.
The algorithm stops, if the $M$ action was used in the output phase.

\paragraph{ALU}
The ALU has $2$ operations, an output operation ($O$) to mark data for output, and a mark operation ($M$), to mark an output as the last, thus $L_B = 2$.
In these four baseline task, these operations signals are only indicating what the algorithm is currently doing, but the ALU has no functionality, i.e., there is no data manipulation.
Hence, the output is just the forwarded input, given by
\begin{align*}
d_t^{out} &= d_t^m \ , \\
c_t^{a} &= c_t^b \ .
\end{align*}

%######################################################################
\section*{Details to the Comparison Methods}
For comparison we used the original Differential Neural Computer~\supercite{graves2016hybrid} (DNC), trained in a supervised setting with gradient descent using the Adam~\supercite{kingma2015adam} optimizer and cross-entropy loss for each step.
The loss is computed based on the correct algorithmic sequences created by the linear layer in the DNC, similar like the fitness function for our architecture, but the cross-entropy loss provides a much richer and localized learning signal.
The DNC is trained for considerable more iterations ($500k$) to counter the pretrained data modules.
It also receives its own output as Input and the data input is managed equally as in our model for each task (see Input module descriptions).
For the repeatCopy task, as in the original implementation, the number of copies $c$ is normalized in the input.
The controller network is a LSTM~\supercite{hochreiter1997long} network with $64$ hidden units except for the Search task, where it has $256$ hidden units to counter the additional data modules.
In the arithmetic task, the modulo $10.000$ operator for intermediate results is replaced with a modulo $10$ operator and the numbers are binary encoded with $4$ bits.
This is done as dealing with decimal input adds an additional challenge and the reduction of the range of the numbers lets the DNC focus on the algorithmic structure of the task, instead of data encoding related issues.
The same bad memories strategy and curriculum schedule as for the NHC are used.
The memory size as well as the number of write and read heads is set to the same values as in the NHC for each task.

As second comparison method, we integrated the DNC into the NHC architecture, named DNC+is+ha.
Therefore, we replaced the NHC algorithmic modules -- Controller, Memory and Bus -- with the original DNC.
To enable this, the information split including the second memory was added to the DNC (+is), and the memory access was changed to hard attentions (+ha), i.e., each head writes and reads one memory location in each step, in contrast to the soft attention and weighted averaged readouts in the DNC.
This is enabled by transforming the computed soft attention heads from the DNC just before memory access into hard attention vectors.
For training this DNC+is+ha model, the exact same learning procedure and parameters were used as for the NHC model.

\printbibliography[segment=\therefsegment, keyword=sup, heading=subbibliography]

\renewcommand\thefigure{Ex.\arabic{figure}}    
\setcounter{figure}{0}    
\figureCurves
\end{refsegment}
\end{document}